\documentclass{article} 
\usepackage{iclr2025_conference,times}
\usepackage{graphicx}
\usepackage{wrapfig}
\usepackage{multirow}
\usepackage{amsfonts}
\usepackage{booktabs}
\usepackage{siunitx}
\usepackage{caption}
\usepackage{amsmath}
\usepackage{natbib}
\usepackage{float}
\usepackage{placeins} 
\usepackage{microtype}

\usepackage{enumitem}

\usepackage{amsmath,amsfonts,bm}

\def\eqref#1{equation~\ref{#1}}

\def\1{\bm{1}}

\DeclareMathAlphabet{\mathsfit}{\encodingdefault}{\sfdefault}{m}{sl}
\SetMathAlphabet{\mathsfit}{bold}{\encodingdefault}{\sfdefault}{bx}{n}

\DeclareMathOperator*{\argmax}{arg\,max}

\usepackage{hyperref}
\usepackage{url}

\title{Why risk matters for protein binder design}

\author{Tudor-Stefan Cotet$^{1,2}$ \& Igor Krawczuk$^{1}$  \\
$^1$Adaptyv, Lausanne, Switzerland \\
$^2$Department of Biosystems Science and Engineering, ETH Zürich, Basel, Switzerland\\
}

\iclrfinalcopy 
\begin{document}

\maketitle

\begin{abstract}
Bayesian optimization (BO) has recently become more prevalent in protein engineering applications and hence has become a fruitful target of benchmarks. However, current BO comparisons often overlook real-world considerations like risk and cost constraints. In this work, we compare 72 model combinations of encodings, surrogate models, and acquisition functions on 11 protein binder fitness landscapes, specifically from this perspective. Drawing from the portfolio optimization literature, we adopt metrics to quantify the cold-start performance relative to a random baseline, to assess the risk of an optimization campaign, and to calculate the overall budget required to reach a fitness threshold. Our results suggest the existence of Pareto-optimal models on the risk-performance axis, the shift of this preference depending on the landscape explored, and the robust correlation between landscape properties such as epistasis with the average and worst-case model performance. They also highlight that rigorous model selection requires substantial computational and statistical efforts.
\end{abstract}

\begin{wrapfigure}{r}{0.45\textwidth}
    \centering
    \includegraphics[width=\linewidth]{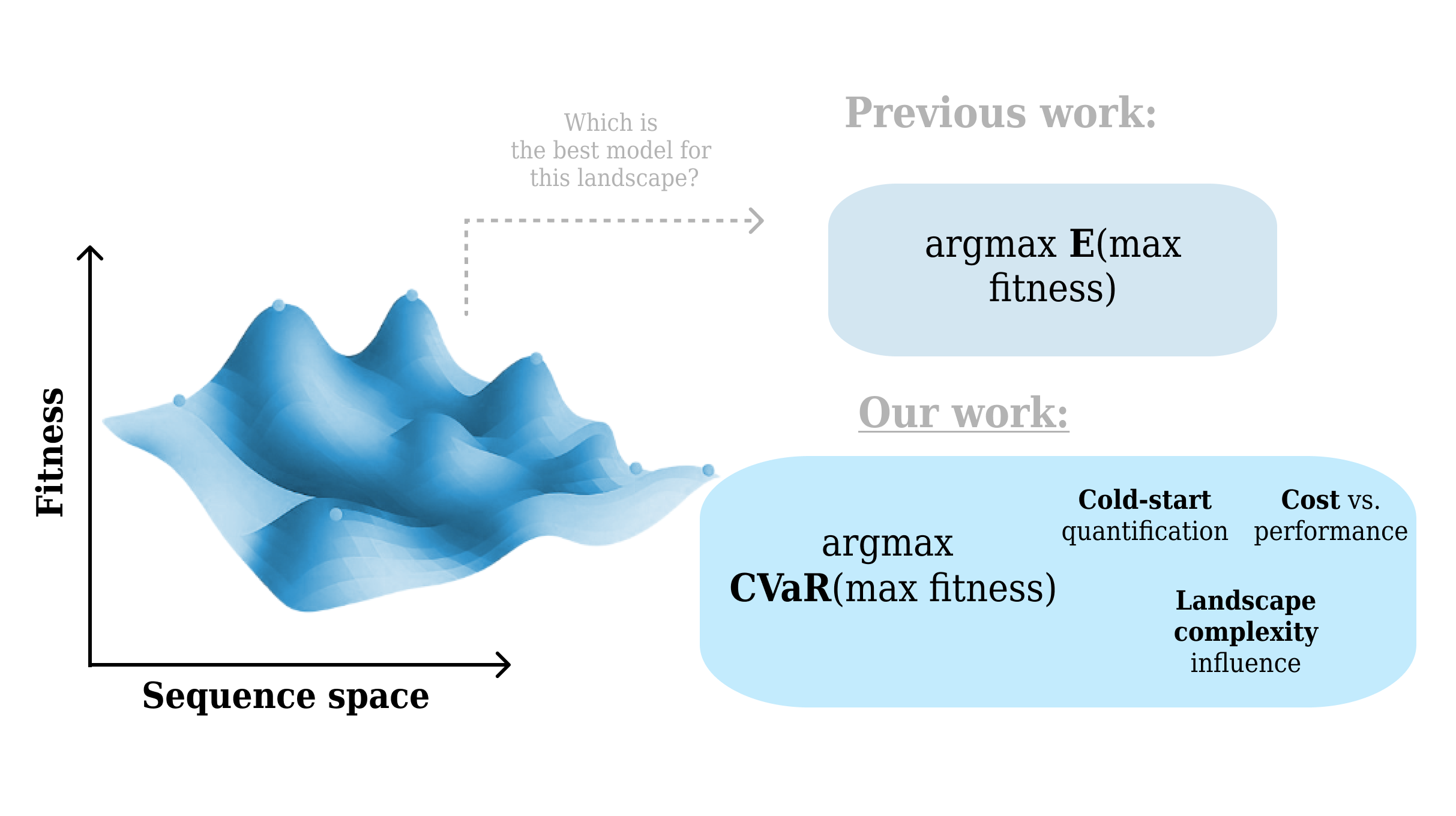}
    \caption{Our additions for protein optimization benchmarking: we consider risks, costs, and performances relative to a random baseline.}
    \label{fig:g1}
\end{wrapfigure}
\section{Introduction}

\textbf{Risk in protein optimization: } In portfolio optimization, the expected value of a portfolio is often an incomplete measure of its desirability: we may not only care about the average of the results, but also their risk \citep{gunjan2023brief}. We can imagine 2 portfolios that both return the same expected value: one of them returns 10\% of average in its worst outcomes and the other returns 90\%. It is pretty clear now which one is more desirable. 

This trade-off also occurs in Bayesian optimization \citep{cakmak2020bayesian, makarova2021risk}. Initialization of parameters or input space, inherent experimental noise, and inadequate exploration can make results highly stochastic \citep{tripp2024diagnosing, wang2022recent}. In protein optimization campaigns, failures (not reaching a target fitness) mean more resources such as time and money will be spent. Thus, when benchmarking models to be used in real campaigns, we should account for and mitigate the risk of failure induced by this stochasticity as much as possible.

\textbf{Previous work:  } Recent work \citep{yang2025active, li2024evaluation, jiang2024rapid} in active learning and Bayesian optimization for proteins has focused on evaluating the performance of the model on landscapes from the ProteinGym \citep{notin2023proteingym} benchmark (although biological test functions are becoming more prevalent, as introduced by \citet{chen2024llms} and \citet{stanton2024closed}). These works then selected the best-performing one to use in an experimental campaign according to the final fitness reached \citep{yang2025active} or fold-change from the starting library \citep{jiang2024rapid}.

\clearpage
To our knowledge, no benchmark for protein optimization explicitly calculated risks, nor addressed budget considerations (as $\$$ amount) or cold-start performance.
This motivates us to study two main questions: 

\begin{enumerate}
  \item \label{question_one} \textbf{How important is risk analysis when benchmarking and selecting Bayesian optimization algorithms for binders?} 
  For this, we investigate the cost savings when ranking models by the CVaR (conditional value at risk) and average performance, and ensure the statistical significance of our results by performing a bootstrap analysis. First, we simulate optimization campaigns for 72 model combinations on 11 binding datasets. We show that, currently, there is no added benefit of a risk-aware ranking, as this is overshadowed by the inherent stochasticity of protein optimization.

  \item \label{question_two} \textbf{Does the fitness landscape influence the average model performance, risks, and costs?} 
  To answer this, we assess the correlation between risk and landscape properties. With the same robust statistical analysis, we show that both the average and CVaR final fitness and costs are greatly influenced by epistasis.
\end{enumerate}

\section{Methods}
\begin{wrapfigure}{r}{0.5\textwidth}
    \centering
    \includegraphics[width=\linewidth]{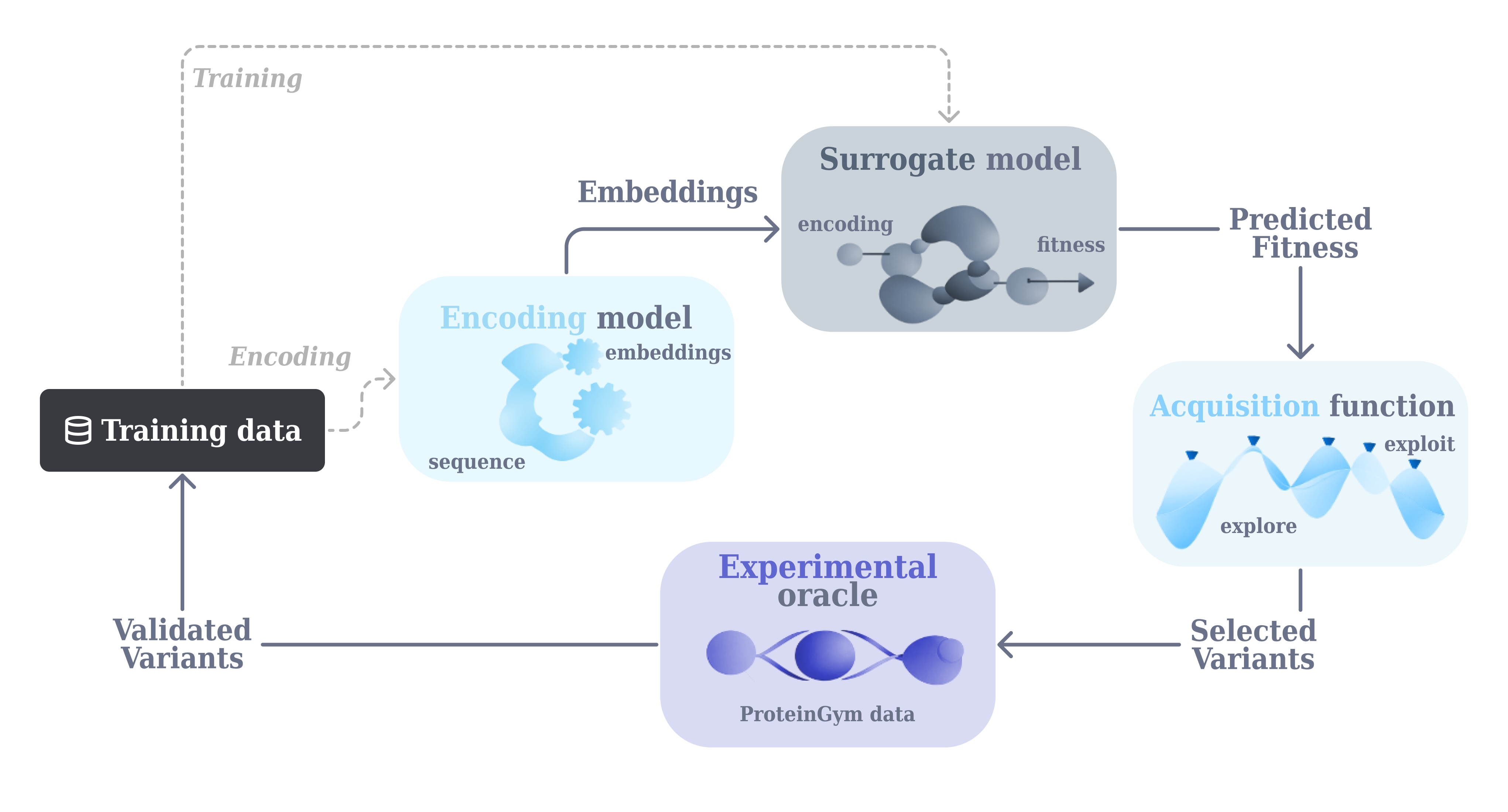}
    \caption{Overview of the standard protein BO loop we are benchmarking.}
    \label{fig:g2}
\end{wrapfigure}

\subsection{Problem definition}
\textbf{Inputs:  } Let $x \in X$ denote a protein sequence of length $L$, where $X$ is the space of all possible sequences composed of amino acids from an alphabet $A$ (typically, $\vert A \vert = 20$). We represent protein sequences using embeddings from a protein language model or with one-hot encodings. Let $e = \phi(x) \in \mathbb{R}^D$ denote the encoding of the sequence $x$, where $\phi: X \to \mathbb{R}^D$ is the embedding function and $D$ is the embedding dimensionality.

\textbf{The black-box function: } In an optimization campaign, our goal is to find the protein sequence $x^*$ that maximizes the unknown fitness function $f$. Since $f$ is expensive to evaluate (i.e., requires performing an assay), Bayesian optimization is commonly used to iteratively propose sequences to evaluate, balancing exploration and exploitation \citep{frazier2018tutorial}. In the context of protein engineering, $f$ represents binding screening experiments or binding affinity measurements, establishing the ground truth fitness values. These can be direct protein-protein binding affinity measurements or more complex proxies, depending on the screening platform. 

\textbf{Surrogates to approximate experimental results: } We use a probabilistic surrogate model to approximate $f$ based on observed data $D_n = \{(x_i, e_i, y_i)\}_{i}^n$, where $y_i = f(x_i) + \epsilon_i$ represents the fitness (i.e., binding affinity) of the oracle and $\epsilon_i \sim N(0, \sigma^2)$ represents the observation noise. Thus, we train a surrogate $\hat{f}$ to predict the fitness value $\hat{y} = \hat{f}(e_i)$.

\textbf{Acquisition functions: } At each iteration, we select the next batch of sequences to evaluate that maximize \citep{wilson2018maximizing} an acquisition function $\alpha(e;\>D_n)$:
\[e_{new\ batch} = \argmax_{e_{candidates} \in \mathbb{R}^D} \alpha(e_{candidates}; D_n)\]

\textbf{Generator and oracle:  } Our pool of candidates $e_{candidates}$ consists of all available sequences in the ground-truth dataset that have not been acquired until the current iteration. Other methods generated all possible mutants, then queried the true label using an L1 distance to the closest in the ground-truth dataset \citep{yang2025active}.

\subsection{Experimental setup and metrics}

\textbf{Datasets and BO components:  } We simulate Bayesian optimization campaigns on 11 binding deep-mutational scanning (DMS) datasets from the ProteinGym repository \citep{notin2023proteingym}, and calculated landscape complexity properties (epistasis, skewness, kurtosis, etc.) for these, described in Appendix \ref{landscape}. We used 6 different uncertainty-aware surrogates: a deep neural network ensemble (ensemble NN), deep neural network with Monte Carlo dropout (dropout NN, \citet{gal2015dropout}), a random forest (RF) model \citep{breiman2001random, louppe2014understanding}, a Gaussian process (GP) model with assumed homoskedastic noise, a deep kernel Gaussian process \citep{wilson2015deep}, and a Bayesian neural network (BNN, \citet{jospin2020hands}). Architecture and hyperparameter optimization details are summarized in Appendix \ref{surrogates}. For the acquisition functions, we implemented greedy top-K, Thompson sampling (TS), expected improvement (EI), and upper-confidence bound (UCB, \citet{srinivas2009gaussian}). We tested one-hot encodings, ESM2-650M, and ESM-3B embeddings mean-pooled \cite{lin2023evolutionary}.

\textbf{Fitness metrics:  } Our main performance metric is the normalized maximum fitness reached at the end of a campaign, commonly used in previous work \citep{li2024evaluation, zhang2024de}, and its conditional value at risk (CVaR) for the worst 10\% of cases. We refer you to \citet{mitra2009risk} and \citet{artzner1999coherent} for an overview of risk measures in financial mathematics. Additional details about our implementation can be found in Appendix \ref{metrics}.

\begin{wrapfigure}[25]{r}{0.5\textwidth}
    \centering
    \vspace{0cm}
    \includegraphics[width=0.5\textwidth]{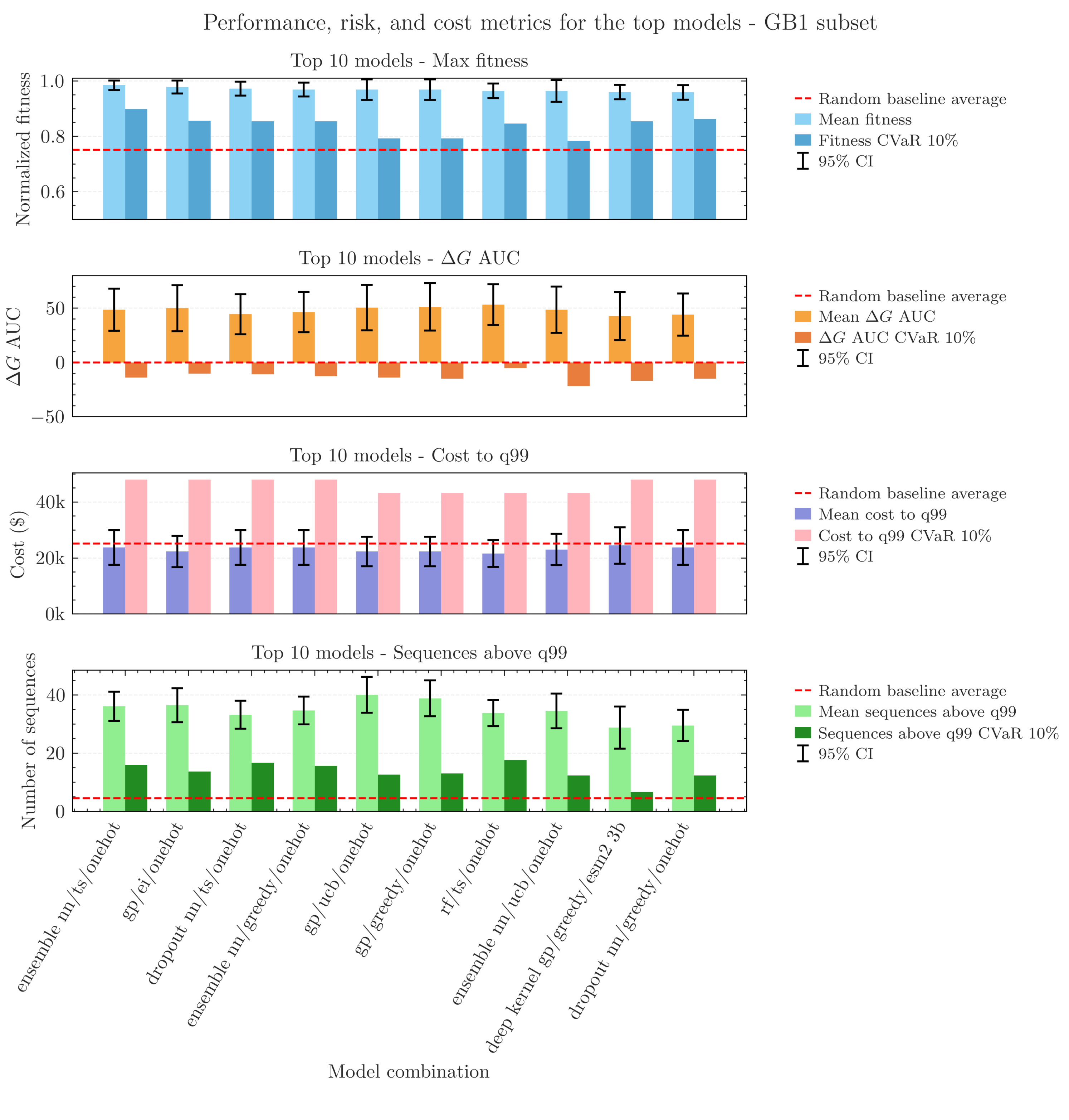}
    \caption{Metrics for the top 10 models ranked by average final fitness for GB1: final fitness reached, $\Delta{G} \> AUC$, cost to 99th percentile of fitness, number of sequences above the 99th percentile threshold acquired.}
    \label{fig:fig1}
\end{wrapfigure}
\FloatBarrier

\textbf{Quantifying cold-starts:  } We have included an additional performance metric accounting for cold-starts, a case in which the model performing similarly to a random baseline in the first steps due to an unadapted surrogate and a small (to none) set of initial training points \citep{poloczek2016warm}. For this, we calculated the difference between the maximum fitness reached by a model versus the random baseline ($\Delta{G}$), further computing its area-under-the-curve ($\Delta{G} \> AUC$). This is a measure of how effectively the model can adapt to and exploit the landscape structure.

\textbf{Quantifying costs:  } Costs were assessed considering an acquisition budget of 384 sequences and a 96 starting pool. Thus, assuming a \$150 price point for testing a single variant \footnote{Based on the cheapest result found when Googling "cheapest binding affinity characterization service wetlab" as of January 2025}, our maximum acquisition budget will be \$57,600 (\$72,000 with the seed library). We have assessed the cost to reach the 99th percentile of fitness for each landscape. This threshold can depend on the priorities of each campaign and we might be more interested in a fold improvement compared to the starting pool for real-life scenarios.

\section{Results}

Returning to our two questions, we first compare CVaR and average-based model rankings (\ref{question_one}), showcase that these agree more on the GB1 landscape, and then determine which landscape property is the best predictor of optimization risk (\ref{question_two}). 
\clearpage

\FloatBarrier
\begin{wrapfigure}{r}{0.4\textwidth} 
    \centering
    \includegraphics[width=0.3\textwidth]{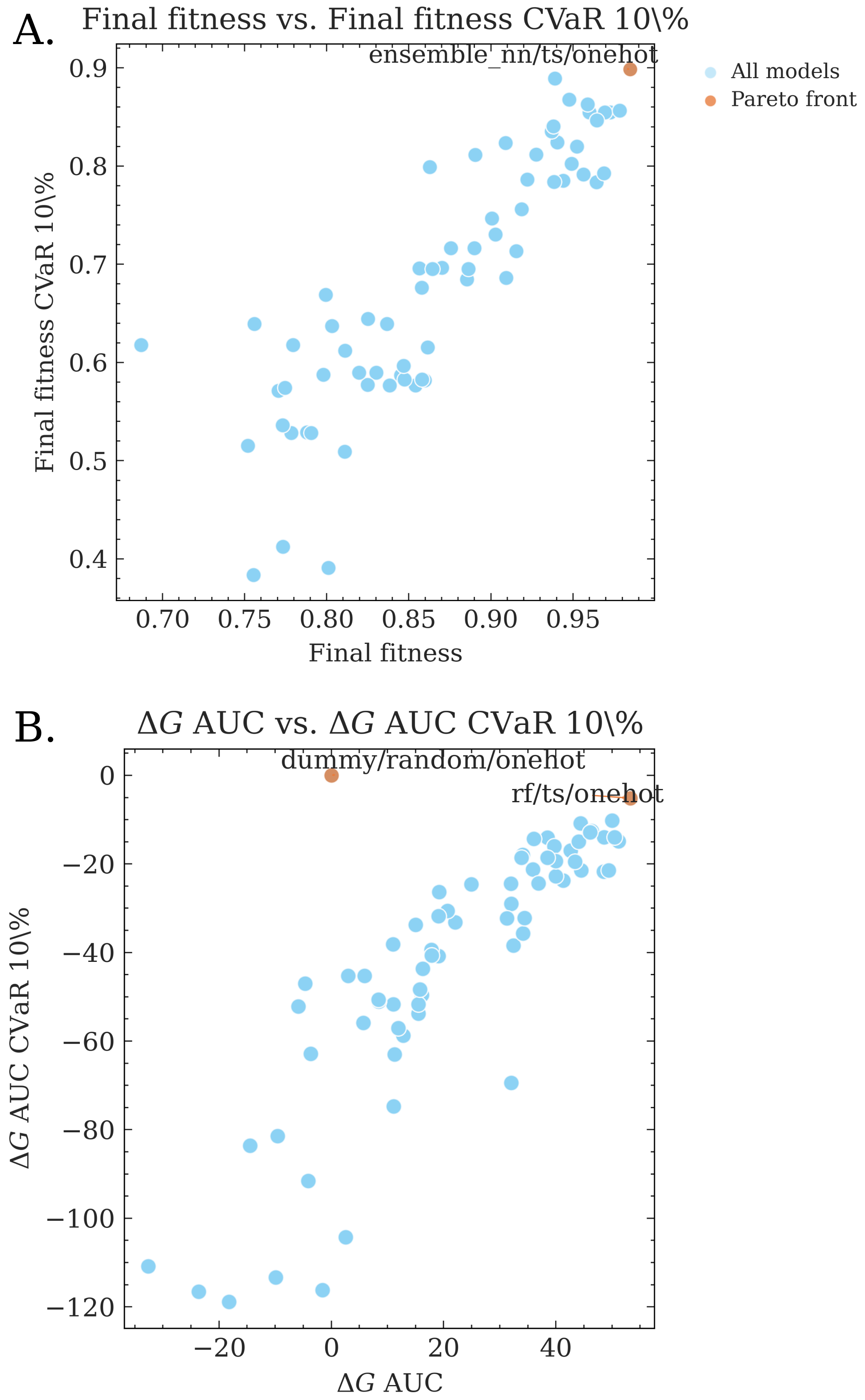} 
    \caption{Pareto frontier when considering the performance-risk axes for the final fitness reached and for the $\Delta{G}\> AUC$ metric.}
    \label{fig:fig2}
\end{wrapfigure}

\textbf{There is a clear risk-performance model preference on the GB1 landscape:  } The GB1 landscape \citep{olson2014comprehensive, wu2016adaptation} has been a workhorse for benchmarking protein optimization algorithms \citep{li2024evaluation, yang2025active, greenman2025benchmarking}. It explores the interaction effects at 4 sites in the IgG-binding domain of protein G (GB1), yielding almost 160,000 total variants. To reduce the computational complexity, we fixed the residue V54 to the wild-type, resulting in 7600 total sequences and 6080 without the subset used for the surrogate hyperparameter search, as detailed in Appendix \ref{surrogates} and \ref{landscape}. 

When ranking models by average final fitness (Supplementary Figure \ref{fig:gb1}), our benchmark recovers the same optimal model identified by the ALDE framework \citep{yang2025active}: DNN ensemble with Thompson sampling and one-hot encodings. In Figure \ref{fig:fig1}, we have summarized several outcome metrics and their associated CVaR values.

This multi-metric analysis reveals a fundamental tension: while GB1's structure allows clear identification of better models under any single metric, different optimization goals can lead to completely different model preferences. For example, the GP/UCB/one-hot model achieves a competitive final fitness, but has a lower CVaR (Figure \ref{fig:fig1}). It still yields the highest number of sequences above the 99th percentile of fitness on average. This suggests that even for well-studied landscapes, model selection should account for the end goals of the optimization campaign. In a typical industrial setting, a risk-aware selection could be more suitable, which we will assess next.

\begin{wraptable}{r}{0.4\textwidth} 
    \centering
    \small
    \begin{tabular}{l r@{}l} 
        \toprule
        {Dataset} & \multicolumn{2}{c}{Kendall $\tau$} \\
        \midrule
        HLA-A & $0.549$ & $^{***}$ \\
        CD19 & $0.252$ & $^{**}$ \\
        CCR5 & $0.212$ & $^{**}$ \\
        ACE2 & $0.681$ & $^{***}$ \\
        CytochromeP4502C9 & $0.568$ & $^{***}$ \\
        Dlg4 PSD95 PDZ3 & $0.606$ & $^{***}$ \\
        KRAS & $0.544$ & $^{***}$ \\
        GB1 subset & $0.667$ & $^{***}$ \\
        YAP1 & $0.769$ & $^{***}$ \\
        SpikeRBD & $0.375$ & $^{***}$ \\
        Gcn4 & $0.149$ & \\
        All datasets & $0.473$ & $^{***}$ \\ 
        \bottomrule
    \end{tabular}
    \caption{Kendall $\tau$ correlation between CVaR and average-based model ranks for each landscape. \textbf{***} denotes a p-value \textless{} 0.001, \textbf{**} for \textless{} 0.01, and \textbf{*} for \textless{} 0.05.}
    \label{tab:fitness_correlation}
\end{wraptable}

The best model is Pareto-dominant when looking at the risk-performance axis (Figure \ref{fig:fig2}). The optimal model differs for $\Delta{G}\>AUC$ (Figure \ref{fig:fig2}). The random forest model might be preferable if we want both to optimize and to have a model adequately learn the structure of a landscape. Several other models become Pareto-optimal when costs are taken into account (Supplementary Figure \ref{fig:suppl - pareto}. These leverage Bayesian NN surrogates and can easily achieve the fitness threshold set, yet they are often not optimizing past it.

\textbf{Variability in landscapes and optimization limits risk-aware model selection: }
Next, we looked at the rank correlations for the 72 models in each landscape (Table \ref{tab:fitness_correlation}). The GB1 subset displays high agreement, while others (e.g., CCR5) show that model preferences vary drastically when accounting for risk in our rankings. We hypothesized that a risk-aware ranking might be more cost-efficient, ideally reducing worst-case costs, and that more complex landscapes would benefit from a risk-aware ranking.

To compare risk-aware and mean-based model selection, we have performed a bootstrap analysis on the 20 runs to better simulate subsequent optimization campaigns, summarized in Appendix \ref{bootstrap}. While naive bootstrapping suggested large potential cost savings (up to \$16,000), a more rigorous testing revealed no statistically significant benefits when accounting for risk into the rankings (Supplementary Figure \ref{fig:suppl-bootstrap}). This suggests we cannot judge based on the simulation whether the risk-aware model choice would actually save money on average or in the worst case. It will require additional seeds and/or k-fold validation, which is a current limitation of our work.

\begin{wrapfigure}{r}{0.4\textwidth} 
    \centering
    \includegraphics[width=0.4\textwidth]{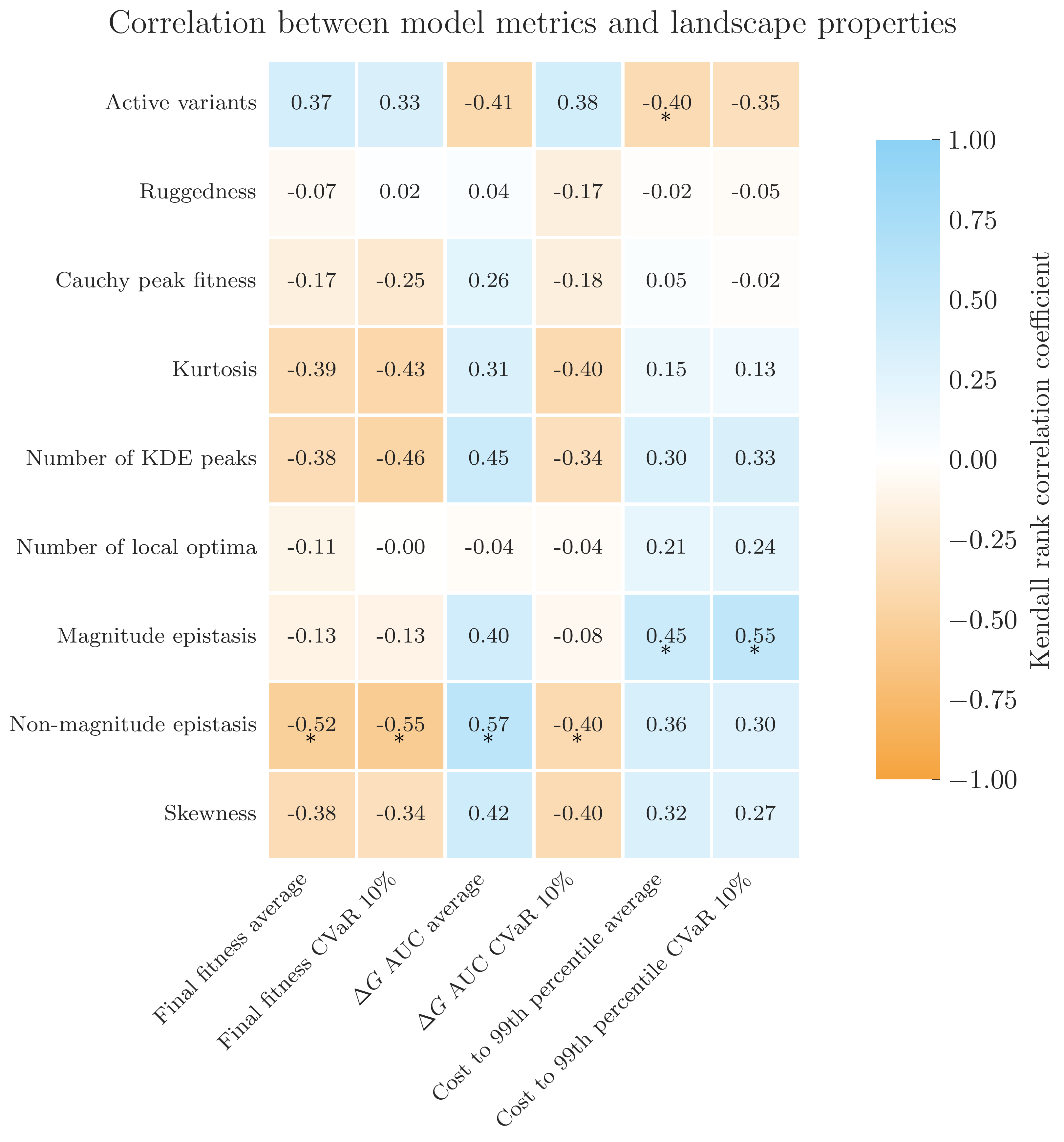} 
    \caption{Kendall $\tau$ correlation coefficients between the average model metrics and landscape properties. \textbf{***} denotes a p-value \textless{} 0.001, \textbf{**} for \textless{} 0.01, and \textbf{*} for \textless{} 0.05.}
    \label{fig:fig3}
\end{wrapfigure}

\textbf{Epistasis is the main driver of increased campaign risks and costs:  } 
Next, we have addressed the second question \ref{question_two} (Figure \ref{fig:fig3}). Bootstrap analysis details are summarized in Appendix \ref{bootstrap}. The final fitness has a strong negative correlation with non-magnitude epistasis (-0.52 average, -0.55 CVaR, statistically significant), while showing moderate negative correlations with kurtosis (-0.39/-0.43), the number of KDE peaks (-0.38/-0.46), and skewness (-0.38/-0.34). This indicates that landscape complexity is directly detrimental to the final performance of all models. The worst 10\% of cases get marginally worse with increased complexity compared to the average. The percentage of active variants shows moderate positive correlations (0.37/0.33), suggesting that an abundance of active variants can stabilize performance.

Several landscape complexity metrics correlate positively with the average $\Delta{G}\>AUC$. Most notably, non-magnitude epistasis (0.57), the number of KDE peaks (0.45), magnitude epistasis (0.40), and skewness (0.42) show substantial positive correlations. However, these metrics generally correlate negatively with the $\Delta{G}\>AUC$ CVaR 10\%, as seen with non-magnitude epistasis (-0.40) and kurtosis (-0.40), suggesting that while distinct landscape topologies are learnable and exploitable on average by models, they can be detrimental when models do not adapt fast enough. Magnitude epistasis shows the strongest positive correlation to the cost (0.45, statistically significant, but a larger confidence interval, Supplementary Figure \ref{fig:suppl-correlconf}) and displays an even stronger correlation in the worst 10\% of cases (0.55). Active variants show consistent negative correlations with costs (-0.40/-0.35), indicating that a higher likelihood of reaching such variants reduces optimization costs. 

Of all correlations, epistasis is the most significant predictor of risk, performance against random, and costs in an optimization campaign, and should be taken into account before starting an optimization campaign.

\section{Conclusion}

In this study, we established the importance of risk quantification and statistically robust benchmarking for protein binder optimization and showed that \textbf{risk-aware BO benchmarking} does indeed give important additional information (\ref{question_one}). We observed reordering effects and breakdown of clear Pareto-dominance when considering risk, costs and performance compared to a random baseline. While rigorous statistical testing did not support the claim of cost savings when changing model selection accordingly, this also remains a tantalizing possibility.

We also showed that there is indeed an \textbf{influence of landscape properties}(\ref{question_two}) and complex landscapes which present an interesting dichotomy: when learned (on average) they yield better-than-random performance, but they correlate with more catastrophic worst-case performance regressions (when learning fails). We find that epistasis is the best predictor of increased risks and costs for the datasets considered.

\textbf{Limitations: }We only evaluated single-objective risk due to the nuances involved in considering multiple objectives. Due to computational constraints, we had to subset some of our datasets, which might change the results. Many more seeds are required to make reliable statistical claims about the impact of risk on model choice. Our model ranking and evaluation approach would benefit from proper k-fold validation using more compute or data for robustness. Finally, we are working with finite datasets where we can only select variants rather than generate new ones. This limits our ability to understand optimization breakdowns in open-ended campaigns.
Subsequent work could implement pre-trained fitness oracles or biological test functions to study these.

\bibliography{iclr2025_conference}
\bibliographystyle{iclr2025_conference}

\clearpage

\appendix
\section{Appendix}
\renewcommand\thefigure{\thesection.\arabic{figure}}    
\setcounter{table}{0}
\renewcommand{\thetable}{\thesection.\arabic{table}} 

\subsection{Surrogate models, acquisition functions, and hyperparameter tuning}
\label{surrogates}

We implemented several standard surrogate models used in previous benchmarks for protein engineering BO \citep{yang2025active} and adapted them to our case (Table \ref{tab:surrogate-architectures-detailed}).     

\begin{table}[H]
\caption{Surrogate model architectures}
\begin{tabular}{l l l}
\toprule
{Surrogate} & Architecture & Activation \\
\midrule
Random forest & Tree ensemble & N/A \\
Deep kernel GP & 3-layer MLP + GP & ReLU \\
Gaussian process & RBF or Matérn kernel and Gaussian likelihood & N/A \\
Bayesian NN & 3-layer BNN & ReLU \\
Dropout NN & 3-layer MLP & ReLU \\
Ensemble NN & 3-layer MLPs & ReLU \\
\bottomrule
\end{tabular}
\label{tab:surrogate-architectures-detailed}
\end{table}

Similarly, we explored 4 commonly used acquisition functions (Table \ref{tab:acquisition-functions}). Subsequent work could further augment these, with protein-specific ones that could improve the optimization process. For example, we could use a classifier to predict the probability of binding along the binding affinity and use it in the acquisition, as done by \citet{rapp2024self}, or augment it with evolutionary or structural priors \citep{frisby2021bayesian}. We believe more work could be done to define principled acquisitions for the binding improvement problem.

\begin{table}[H]
\caption{Acquisition functions implemented in our benchmark. $\mu(x)$ and $\sigma(x)$ are the predicted mean and standard deviation from the surrogate model, $f^*$ is the current best-observed value, $\Phi$ and $\phi$ are the CDF and PDF of the standard normal distribution. We did not perform a hyperparameter search for the acquisition function.}
\begin{tabular}{l l l}
\toprule
{Acquisition} & Formula & Parameters \\
\midrule
Expected improvement & $\mathbb{E}[\max(f(x) - f^* - \xi, 0)]$ & $\xi = 0.01$ \\
              & $= ({\mu(x) - f^* - \xi})\Phi(z) + \sigma(x)\phi(z)$ & \\
              & where $z = \frac{\mu(x) - f^* - \xi}{\sigma(x)}$ & \\
\midrule
Upper confidence bound & $\mu(x) + \beta\sigma(x)$ & $\beta = 2.0$ \\
\midrule
Thompson sampling & $f(x) \sim \mathcal{N}(\mu(x), \sigma^2(x))$ & None \\
\midrule
Greedy & $\mu(x)$ & None \\
\bottomrule
\end{tabular}
\label{tab:acquisition-functions}
\end{table}

For our optimization campaign, we conducted a comprehensive evaluation of model combinations to determine their peak performance and ensure our model comparisons as robust. We implemented a grid search across multiple hyperparameters, testing each parameter configuration on individual landscapes. In each landscape, we allocated 20\% of the sequences for hyperparameter optimization (15\% training, 5\% testing), while reserving the remaining 80\% for the actual simulated campaign. We chose a random split over a homology or a high-low activity one \citep{dallago2022flip} because we wanted to maintain the inherent fitness distribution and preserve the landscape topology. We selected our grid samples by consulting the literature, both for previous protein BO campaigns, the original surrogate implementations, and standardized implementations from libraries like GPyTorch (e.g., the number of Monte-Carlo samples or dropout rate as highlighted by \citet{gal2015dropout}).

We used a schedule-free implementation of the Adam optimizer for both the hyperparameter tuning and simulated campaigns, as initially described by \citet{defazio2024road}.

\begin{table}[H]
\caption{Hyperparameter search space for each surrogate model. Common parameters across neural models: no. epochs = 100, batch size = 32, Monte Carlo samples = 30 (where applicable).}
\label{tab:surrogate-hyperparameters}
\centering
\begin{tabular}{l l}
\toprule
{Surrogate} & Hyperparameter search space \\
\midrule
Random forest & no. estimators $\in \{10, 50, 100, 200\}$ \\
             & max depth $\in \{None, 10\}$ \\
\midrule
Deep kernel GP & hidden dimension = 128 \\
              & learning\_rate $\in \{10^{-4}, 5{\times}10^{-4}, 10^{-3}, 5{\times}10^{-3}, 10^{-2}, 5{\times}10^{-2}, 10^{-1}\}$ \\
              & kernel\_type $\in \{$RBF, Matérn$\}$ \\
\midrule
Gaussian process & kernel type $\in \{$RBF, Matérn$\}$ \\
                & learning rate $\in \{10^{-4}, 5{\times}10^{-4}, 10^{-3}, 5{\times}10^{-3}, 10^{-2}, 5{\times}10^{-2}, 10^{-1}\}$ \\
\midrule
Bayesian NN & hidden dimension = 128 \\
           & learning rate $\in \{10^{-4}, 5{\times}10^{-4}, 10^{-3}, 5{\times}10^{-3}, 10^{-2}, 5{\times}10^{-2}, 10^{-1}\}$ \\
           & KL weight = 1.0 \\
\midrule
Dropout NN & hidden dimension = 128 \\
          & learning rate $\in \{10^{-4}, 5{\times}10^{-4}, 10^{-3}, 5{\times}10^{-3}, 10^{-2}, 5{\times}10^{-2}, 10^{-1}\}$ \\
          & dropout rate = 0.1 \\
\midrule
Ensemble NN & hidden dimension = 128 \\
           & learning rate $\in \{10^{-4}, 5{\times}10^{-4}, 10^{-3}, 5{\times}10^{-3}, 10^{-2}, 5{\times}10^{-2}, 10^{-1}\}$ \\
           & no. estimators = 5 \\
\bottomrule
\end{tabular}
\label{tab:surrogate-hyperparameters}
\end{table}

\subsection{Landscape property calculations}
\label{landscape}

We calculated the properties of our landscapes following the methodology established by \citet{li2024evaluation}, with our main difference in using the Otsu method to determine the threshold for active variants. These are summarized in the table below. N represents the total number of variants. Ruggedness (Rugged.) quantifies local structure complexity. Peak fitness shows the Cauchy distribution peak location. Kurt. (kurtosis) and Skew (skewness) describe the fitness distribution shape. KDE peaks indicate the number of modes in the kernel density estimation. Optima shows the count of local fitness maxima. Magnitude (Mag.) and non-magnitude (Non-mag.) epistasis percentages quantify pairwise interaction types.

\begin{table}[H]
\small
\caption{Landscape properties.}
\label{tab:landscape-properties}
\begin{tabular}{l r@{.}l r@{.}l r r@{.}l r@{.}l r@{.}l r r r@{.}l r@{.}l r@{.}l} 
\toprule
Dataset & \multicolumn{2}{c}{Active\%} & \multicolumn{2}{c}{Thresh.} & N & \multicolumn{2}{c}{Rugged.} & \multicolumn{2}{c}{Peak} & \multicolumn{2}{c}{Kurt.} & KDE & Local optima & \multicolumn{2}{c}{Mag.} & \multicolumn{2}{c}{Non-mag.} & \multicolumn{2}{c}{Skew} \\
& \multicolumn{2}{c}{(Otsu)} & \multicolumn{2}{c}{(Otsu)} & & & & \multicolumn{2}{c}{fitness} & & & peaks & & \multicolumn{2}{c}{epist.} & \multicolumn{2}{c}{epist.} & & \\
\midrule
KRAS & 57&59 & -0&56 & 19899 & 0&21 & -0&38 & -1&16 & 3 & 576 & 25&31 & 74&69 & -0&36 \\
GB1 subset & 3&82 & 1&32 & 6080 & 1&73 & 0&00 & 36&53 & 14 & 7 & 6&71 & 93&29 & 5&61 \\
SpikeRBD & 82&40 & -2&08 & 3042 & 0&90 & -0&19 & 1&47 & 2 & 186 & 0&00 & 0&00 & -1&64 \\
CytP4502C9 & 51&01 & 0&53 & 4914 & 0&40 & 0&54 & -1&41 & 2 & 465 & 0&00 & 0&00 & 0&07 \\
CCR5 & 75&41 & -0&37 & 4910 & 0&74 & -0&02 & 5&18 & 4 & 323 & 0&00 & 0&00 & -1&05 \\
CD19 & 17&65 & 0&30 & 3009 & 4&24 & -2&02 & 1&82 & 2 & 269 & 0&00 & 0&00 & 1&01 \\
Gcn4 & 45&75 & 1&32 & 2111 & 0&30 & 1&29 & 16&89 & 6 & 47 & 0&00 & 100&00 & 1&02 \\
ACE2 & 63&07 & -0&66 & 1779 & 1&40 & -0&25 & 0&72 & 2 & 117 & 0&00 & 0&00 & 0&19 \\
HLA-A & 65&73 & -0&18 & 2676 & 1&35 & 0&22 & 0&91 & 2 & 178 & 0&00 & 0&00 & -0&54 \\
YAP1 & 21&45 & 1&29 & 8060 & 1&21 & 0&54 & 12&11 & 9 & 926 & 5&49 & 94&51 & 2&52 \\
Dlg4 PDZ3 & 87&63 & -0&59 & 1261 & 0&29 & 0&01 & 3&48 & 3 & 83 & 0&00 & 0&00 & -2&03 \\
\bottomrule
\end{tabular}
\end{table}

\clearpage
\subsection{Optimization risk, performance, and cost calculations}
\label{metrics}
\textbf{Cold-start quantification using the $\Delta{G} \> AUC$: } We define the optimization strategy $O_i$ as the combination of surrogate, acquisition, encoding, loss, and kernel for a given choice of these parameters $i$. We set the initial starting pool of variants as $n_{init}^s$ for the initialization (seed) $s$ and $n_{k}^{s}$ as the current acquired variants at iteration $k$. In total, we set a fixed budget $n_{budget} = n_{init} + b\>*\>K$, where $b$ is the batch size and $K$ is the number of optimization cycles.

We define $G_k^{s}$ as the payoff (reward, performance) achieved after conducting the optimization at iteration $k$ accounting for $O_i$, $n_{init}^{s}$, and $n_{k}^{s}$. For single-objective optimization, $G_{k}^{s}$ will be:

$$
G^{s}_{k}(O_i, n^s_{\text{init}}, n^s_{k}) = \max_{x^{s}_{k} \in X} f(x^{s}_{k}) \mid O_i, n^s_{\text{init}}, n^s_k)
$$

Where $x_k^s$ contains all variants acquired including iteration $k$ for seed $s$.

Thus, we are continuously aware of our budget and variants acquired at each iteration when computing the payoff. We have defined payoff as the maximum acquired binding affinity up until the $k$th iteration, but other metrics such as cumulative regret could be used.

We want to calculate the $\Delta G$ difference in optimum values for a given model choice $O_i$ and a simple baseline $O_0$ for the early stages of optimization, for a given seed. The simple baseline will be a random search in our case, which mimics the approximate behaviour of a Bayesian optimization algorithm with uniformly distributed priors in the early stages of optimization to quantify cold starts.

$$
\Delta G^s_k=G^s_k(O_i,n^s_{init},n^s_k)-G^s_k(O_0,n^s_{init},n^s_k)
$$

Thus, cold starts are represented by a small $\Delta G_k^s$: the model needs to learn more about the data to suggest optimal value variants, as it resembles a random search initially. A large $\Delta G_k^s$ suggests effective utilization of prior information, leading to better performance.

Furthermore, we can average $\Delta{G}^s_k$ across all seeds, with $\Delta{G_k} = \mathbb E[\Delta{G_k^s} \mid s \in S]$. Furthermore, we will compute the Area Under the Curve (AUC) to obtain $\Delta{G}^s_{1:K}\>AUC$ - the cumulative model’s performance considering all time steps (k from 1 to K). This ensures we rank highly models that do not necessarily converge to the optimum, but achieve a steady improvement throughout all iterations (especially if we want to stop an optimization earlier if we are content with the results).

\textbf{Risk metrics for ranking: } 
We are interested in quantifying the performance in worst-case scenarios, and thus implemented risk measures from financial mathematics - VaR (value at risk) and CVaR (Conditional Value at Risk, also known as the Expected Shortfall, ES). \citep{artzner1999coherent, cakmak2020bayesian}.

$$
\text{VaR}_\alpha(X) = \inf\{x \in \mathbb{R} : P(X \leq x) \geq \alpha\}
$$

$$
\text{CVaR}_\alpha(X) = \mathbb{E}[X \mid X \leq \text{VaR}_\alpha(X)]
$$

In our benchmark, we have used the CVaR/ES with an $\alpha = 0.1$ calculated for the final fitness performance, performance relative to baseline, and costs,  determining the average of payoffs that do not reach the VaR threshold (fitness in the worst 10\% of cases). To make mathematical notation less abstract, a large ES means that even in the worst 10\% of initializations, a model will still yield a high payoff throughout all iterations (AUC), and will outperform a random baseline (positive AUC). An ES close to the maximum AUC means the model is robustly parsing the fitness landscape, irrespective of its initialization.

\clearpage
\subsection{Bootstrap analysis}
\label{bootstrap}
Two bootstrap analyses were performed to address the limited number of runs (seeds) sampled.

\subsubsection{Comparing mean versus CVaR-based model ranking}
\begin{figure}[!ht]
    \centering
    \includegraphics[width=1.1\textwidth]{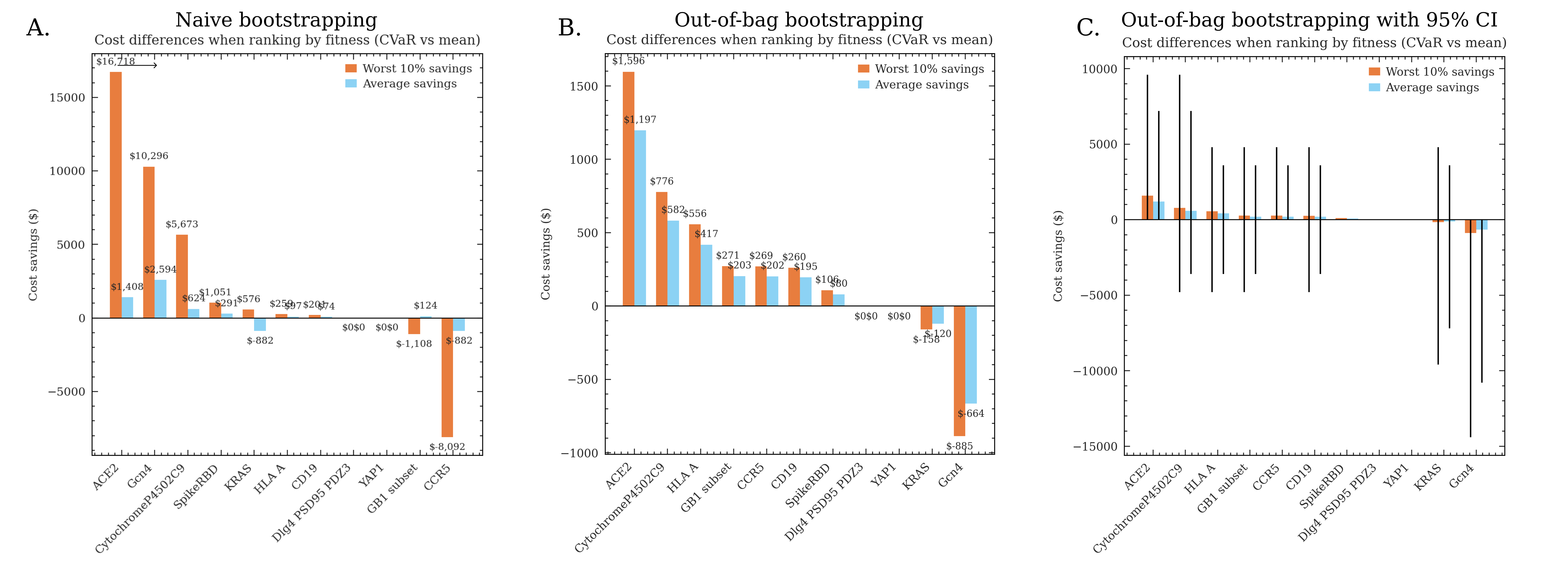}
    \caption{Comparison of different bootstrapping and evaluation techniques to establish the cost savings between CVaR-based and mean-based model ranking.}
    \label{fig:suppl-bootstrap}
\end{figure}

We performed a bootstrap analysis with increasing stringency to assess the statistical significance of the mean and CVaR model rankings and their associated campaign cost differences. For the naive bootstrap, we sampled with replacement from the available optimization runs (20 seeds) for each dataset-model pair. We then ranked the top models for each bootstrap sample using either the average fitness over the seeds or the CVaR, selected the best one for each strategy, and then compared the corresponding cost difference to reach the 99th percentile fitness between the CVaR and mean strategies. We did not split the runs into ranking and evaluation ones. This yielded some impressive (albeit overestimating benefits due to data leakage) results: CVaR-based ranking saved upwards of 25\% of the optimization campaign, improved average outcomes, and had a negative trend only for a single complex landscape with highly variable campaign outcomes (Figure \ref{fig:suppl-bootstrap} A).

Next, we implemented an out-of-bag bootstrap strategy: 80\% of seeds were used for model ranking (16 seeds) and the remaining were reserved for evaluation. We then generated all possible combinations of ranking and evaluation seeds, evaluated the average and worst-case (of the evaluation seeds) cost savings, and calculated confidence intervals using the percentile method (Figure \ref{fig:suppl-bootstrap} B and C). While naive bootstrapping suggested substantial benefits from risk-aware selection, more rigorous statistical testing revealed these differences were not significant enough to justify favouring one ranking method over the other. We tried to address the lower number of runs per model via the bootstrapping approach, yet this is still a limitation of our work.

\subsubsection{Establishing the confidence of correlations between landscape properties and model performances}
To account for the limited number of runs per model type, we bootstrapped (sampling with replacement) the seeds, then calculated the CVaR and average model performance for our 3 metrics, averaged them for each landscape, and computed the Kendall $\tau$ correlation scores between the average model performance and landscape properties. This was done for 1000 bootstrap samples and confidence intervals were calculated with the percentile method. We observed similar confidence intervals for increasing bootstrap sample sizes. Our results are summarized below. Overall, the correlation coefficients between final fitness and properties are confidently established, yet they vary more when considering the cost metrics. We assume this is due to some sequences above the 99th threshold being acquired earlier simply by chance, even if the model cannot effectively exploit this fitness region.

\begin{figure}[H]
    \centering
    \includegraphics[width=0.75\textwidth]{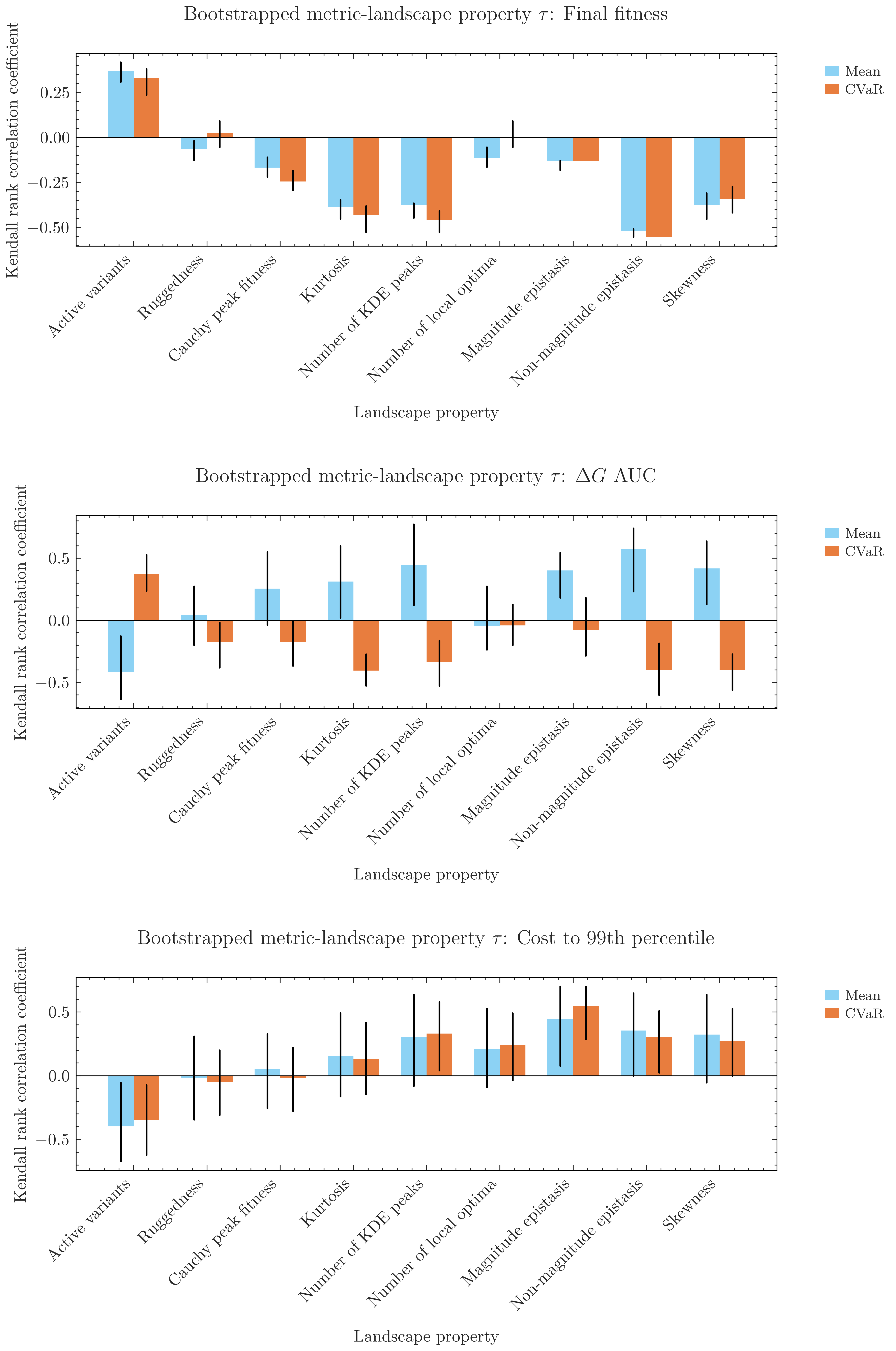}
    \caption{Kendall $\tau$ values following the bootstrap analysis between the average model performance or risk and several landscape properties. Error bars indicate the 95\% confidence intervals.}
    \label{fig:suppl-correlconf}
\end{figure}

\clearpage
\subsection{Additional rank correlations}
We have highlighted additional Kendall $\tau$ correlations between risk and average-based rankings for the $\Delta{G}\> AUC$ and cost metrics (Table \ref{tab:agreement-metric}). Seemingly, models tend to require similar numbers of sequences regardless of ranking method to reach our fitness threshold, indicating it might either be too strict or easy to reach. This showcases the difficulty of setting \textit{a priori} absolute optimization goals. Subsequent analyses could set an achievable fold-change goal instead. When considering the performance relative to a baseline some models that perform well on average struggle in worst cases (e.g., on the CCR5 landscape), suggesting the landscape may have regions where models fail to distinguish themselves from random search. 

\begin{table}[H]
    \caption{Rank agreement (as Kendall $\tau$ correlation) between CVaR and average-based model rankings. Statistical significance: $^{*}p<0.05$, $^{**}p<0.01$  $^{***}p<0.001$.}
    \label{tab:agreement-metric}

    \begin{tabular}{l r@{}l r@{}l} \toprule
        {Dataset} & \multicolumn{2}{c}{$\Delta$G AUC rank} & \multicolumn{2}{c}{Cost to 99th percentile} \\
        & \multicolumn{2}{c}{correlation} & \multicolumn{2}{c}{rank correlation} \\ \midrule
        HLA-A & $0.519$ & $^{***}$ & $0.937$ & $^{***}$ \\
        CD19 & $0.487$ & $^{***}$ & $0.919$ & $^{***}$ \\
        CCR5 & $0.295$ & $^{***}$ & $0.933$ & $^{***}$ \\
        ACE2 & $0.677$ & $^{***}$ & $1.000$ & $^{***}$ \\
        CytochromeP4502C9 & $0.591$ & $^{***}$ & $0.759$ & $^{***}$ \\
        Dlg4 PSD95 PDZ3 & $0.588$ & $^{***}$ & $0.971$ & $^{***}$ \\
        KRAS & $0.546$ & $^{***}$ & $0.703$ & $^{***}$ \\
        GB1 subset & $0.720$ & $^{***}$ & $0.834$ & $^{***}$ \\
        YAP1 & $0.677$ & $^{***}$ & $0.581$ & $^{***}$ \\
        SpikeRBD & $0.475$ & $^{***}$ & $0.976$ & $^{***}$ \\
        Gcn4 & $0.369$ & $^{***}$ & $0.913$ & $^{***}$ \\
        All datasets & $0.529$ & $^{***}$ & $0.816$ & $^{***}$ \\ \bottomrule
    \end{tabular}
\end{table}

\subsection{Rank agreement versus landscape property correlations}
We further analysed which landscape properties influence the rank correlation between CVaR and average-based methods. We did not observe any statistically significant results.

\begin{table}[H]
    \caption{Correlation between landscape properties and rank agreement between average and worst-case (CVaR) metrics. Statistical significance: $^{*}p<0.05$, $^{**}p<0.01$  $^{***}p<0.001$.}
    \label{tab:landscape-agreement-correlation}
    \begin{tabular}{l r r r} \toprule
        \multirow{2}{*}{Landscape property} & Final fitness & $\Delta$G AUC & Cost to 99th \\
        & rank correlation & rank correlation & percentile correlation \\ \midrule
        Active variants \% (Otsu) & -0.422 & 0.000 & -0.467 \\
        Ruggedness & -0.156 & -0.135 & 0.156 \\
        Cauchy peak fitness & -0.180 & -0.432 & 0.225 \\
        Kurtosis & 0.022 & -0.360 & -0.333 \\
        Number of KDE peaks & 0.227 & 0.102 & 0.025 \\
        Number of local optima & -0.244 & 0.000 & 0.244 \\
        Magnitude epistasis & 0.243 & -0.062 & 0.243 \\
        Non-magnitude epistasis & 0.109 & -0.028 & 0.218 \\
        Skewness & 0.067 & 0.360 & -0.289 \\ \bottomrule
    \end{tabular}
\end{table}

\clearpage
\subsection{Optimization curves for the top models across different landscapes}

\subsubsection{GB1 top models}
\begin{figure}[H]
    \centering
    \includegraphics[width=1\textwidth]{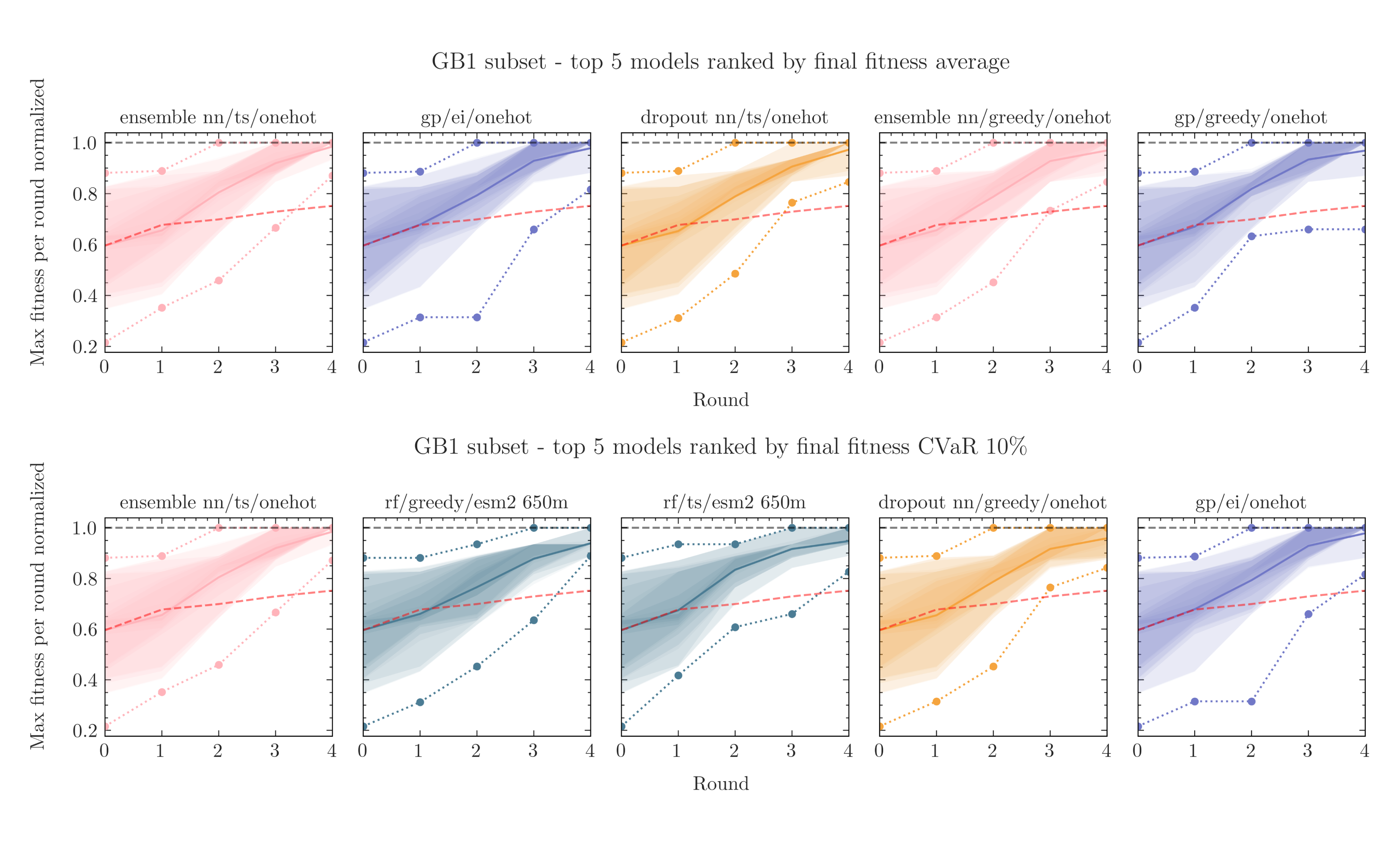}
    \caption{Top 5 models for the average and CVaR-based selection on the final fitness reached - GB1 subset dataset.}
    \label{fig:gb1}
\end{figure}

\subsubsection{CCR5 top models}
\begin{figure}[H]
    \centering
    \includegraphics[width=1\textwidth]{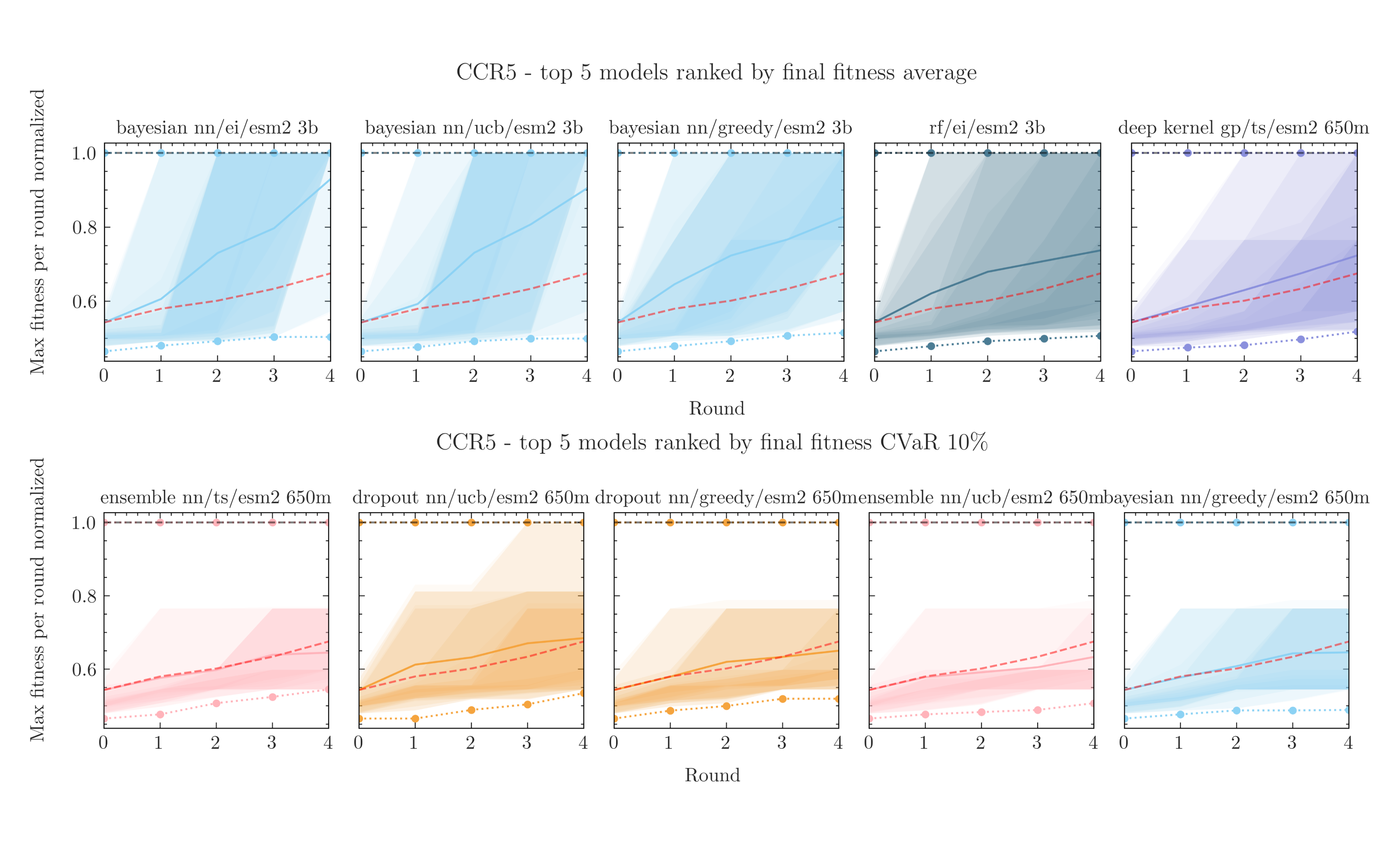}
    \caption{Top 5 models for the average and CVaR-based selection on the final fitness reached - CCR5 dataset.}
    \label{fig:ccr5}
\end{figure}

\subsubsection{CD19 top models}
\begin{figure}[H]
    \centering
    \includegraphics[width=1\textwidth]{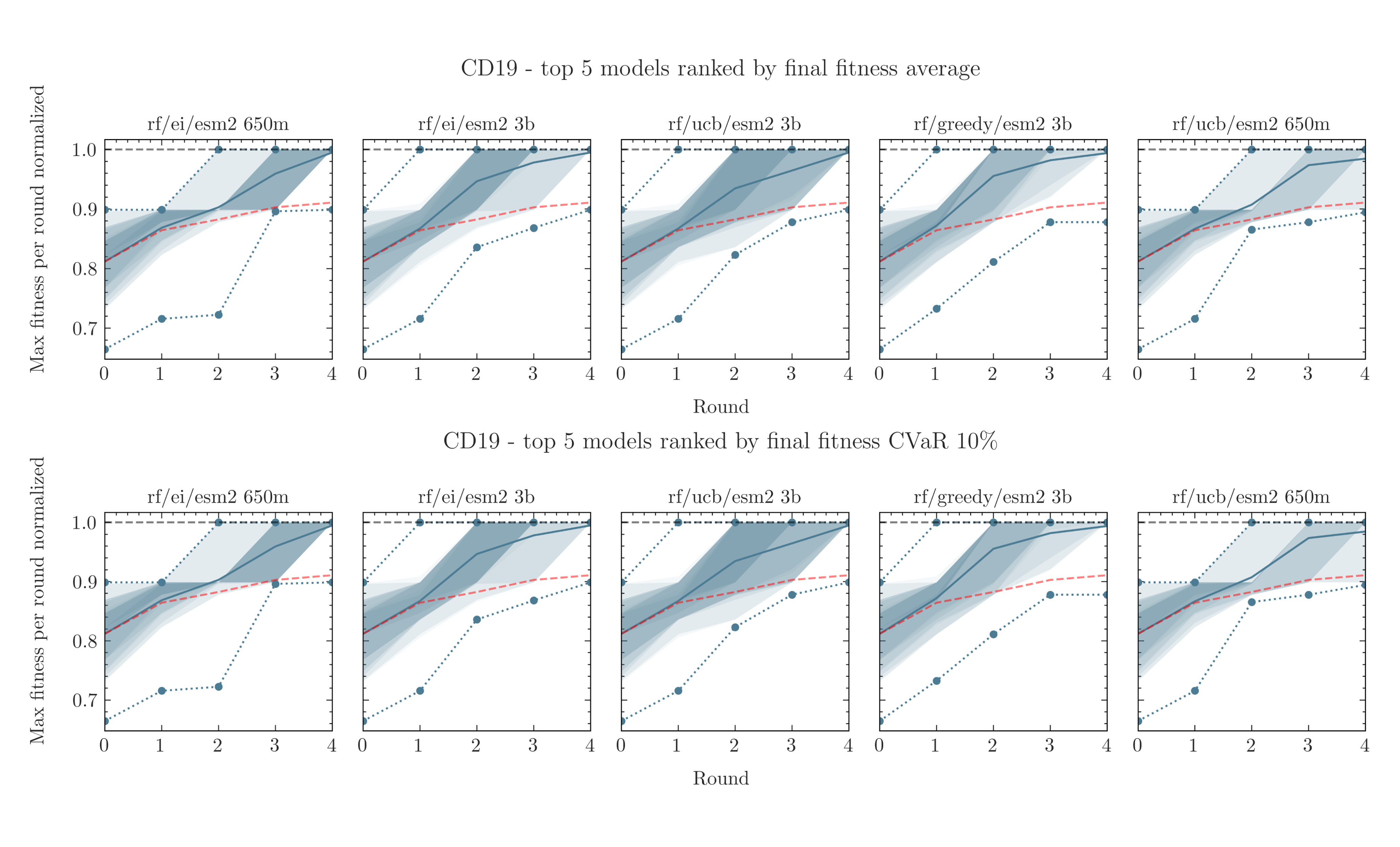}
    \caption{Top 5 models for the average and CVaR-based selection on the final fitness reached - CD19 dataset.}
    \label{fig:cd19}
\end{figure}

\subsubsection{HLA-A top models}
\begin{figure}[H]
    \centering
    \includegraphics[width=1\textwidth]{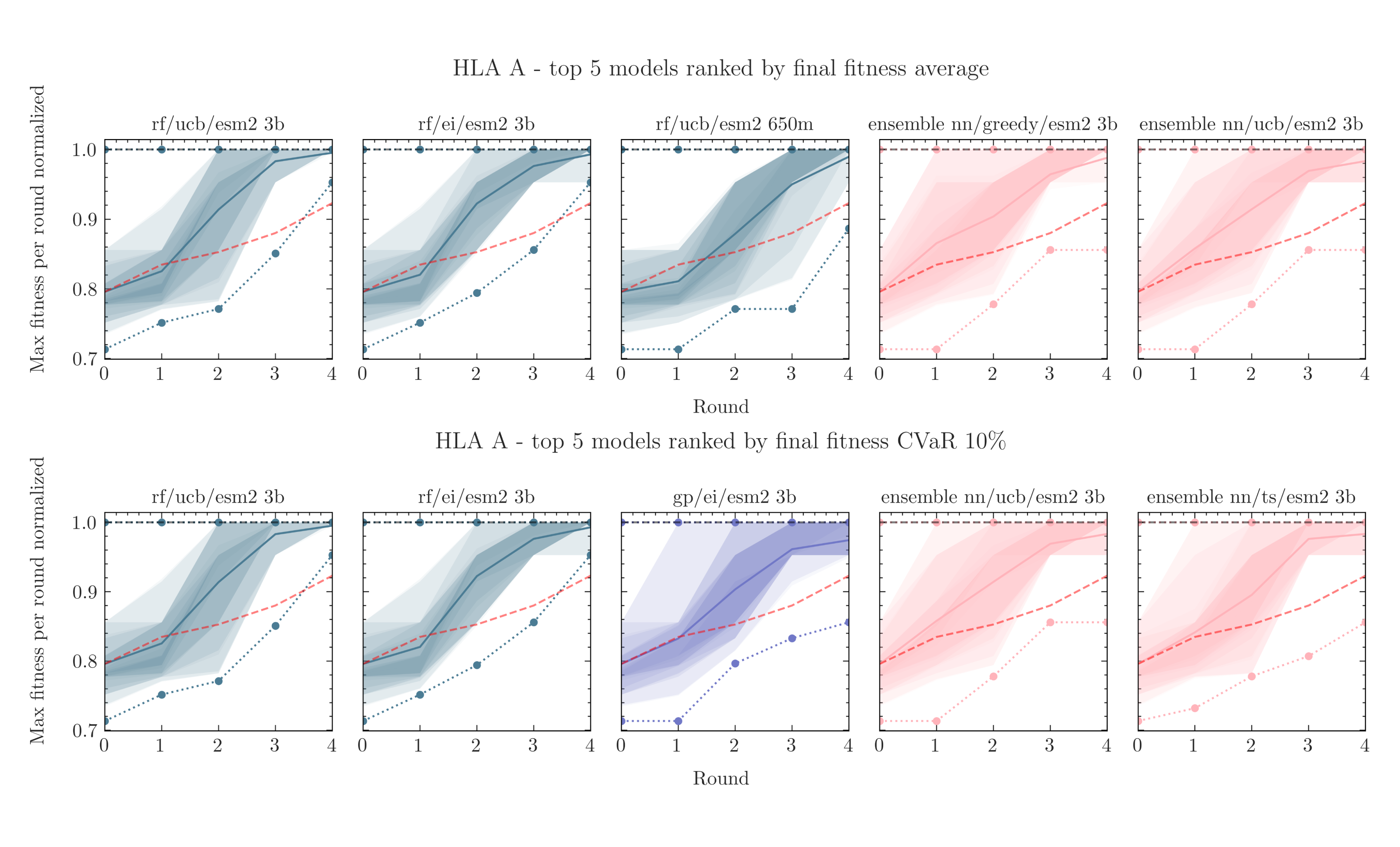}
    \caption{Top 5 models for the average and CVaR-based selection on the final fitness reached - HLA-A dataset.}
    \label{fig:hlaa}
\end{figure}

\subsubsection{Cytochrome P4502C9 top models}
\begin{figure}[H]
    \centering
    \includegraphics[width=1\textwidth]{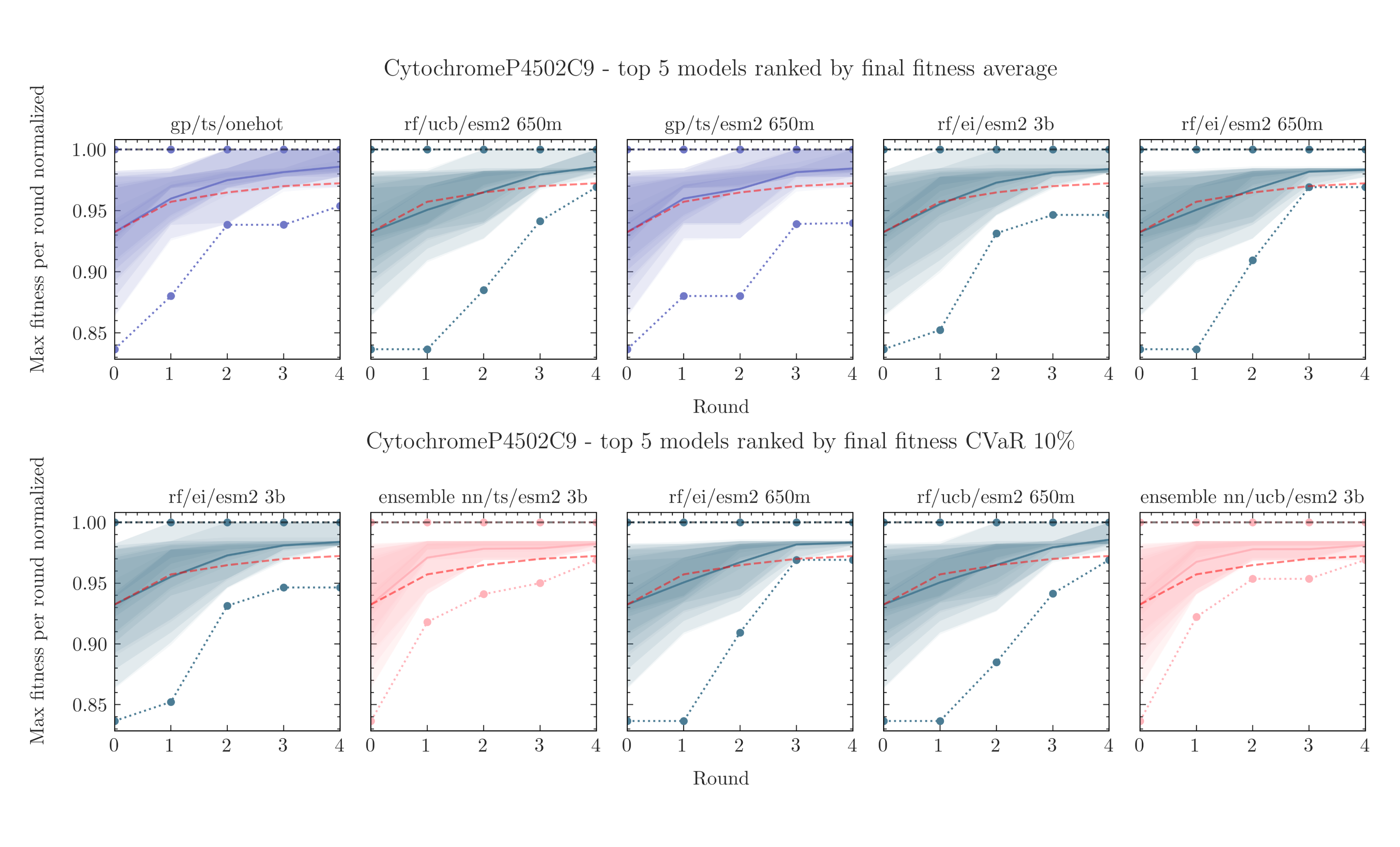}
    \caption{Top 5 models for the average and CVaR-based selection on the final fitness reached - Cytochrome P4502C9 dataset.}
    \label{fig:cytochrome}
\end{figure}

\subsubsection{KRAS1 top models}
\begin{figure}[H]
    \centering
    \includegraphics[width=1\textwidth]{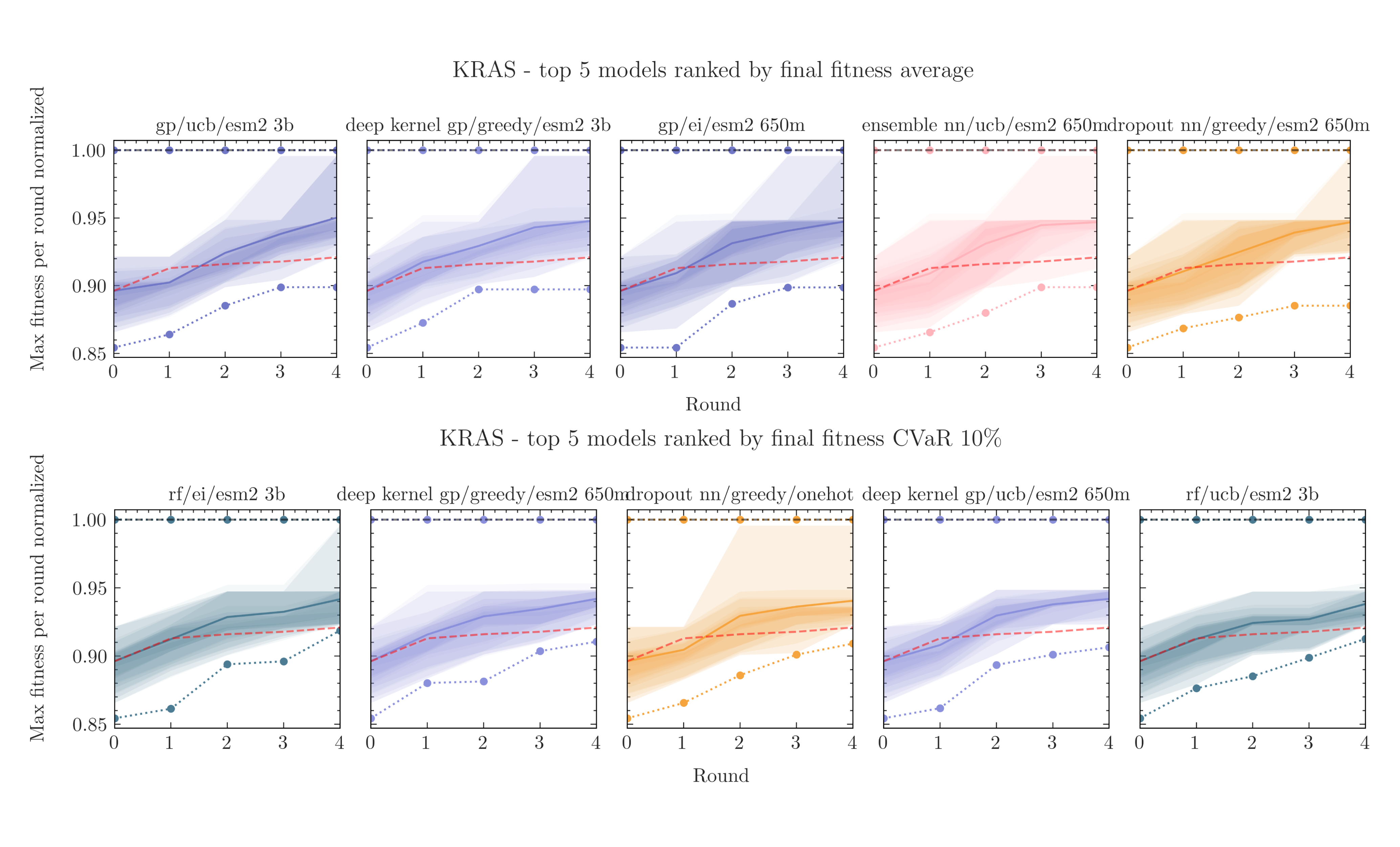}
    \caption{Top 5 models for the average and CVaR-based selection on the final fitness reached - KRAS1 dataset.}
    \label{fig:hlaa}
\end{figure}

\subsubsection{SpikeRBD top models}
\begin{figure}[H]
    \centering
    \includegraphics[width=1\textwidth]{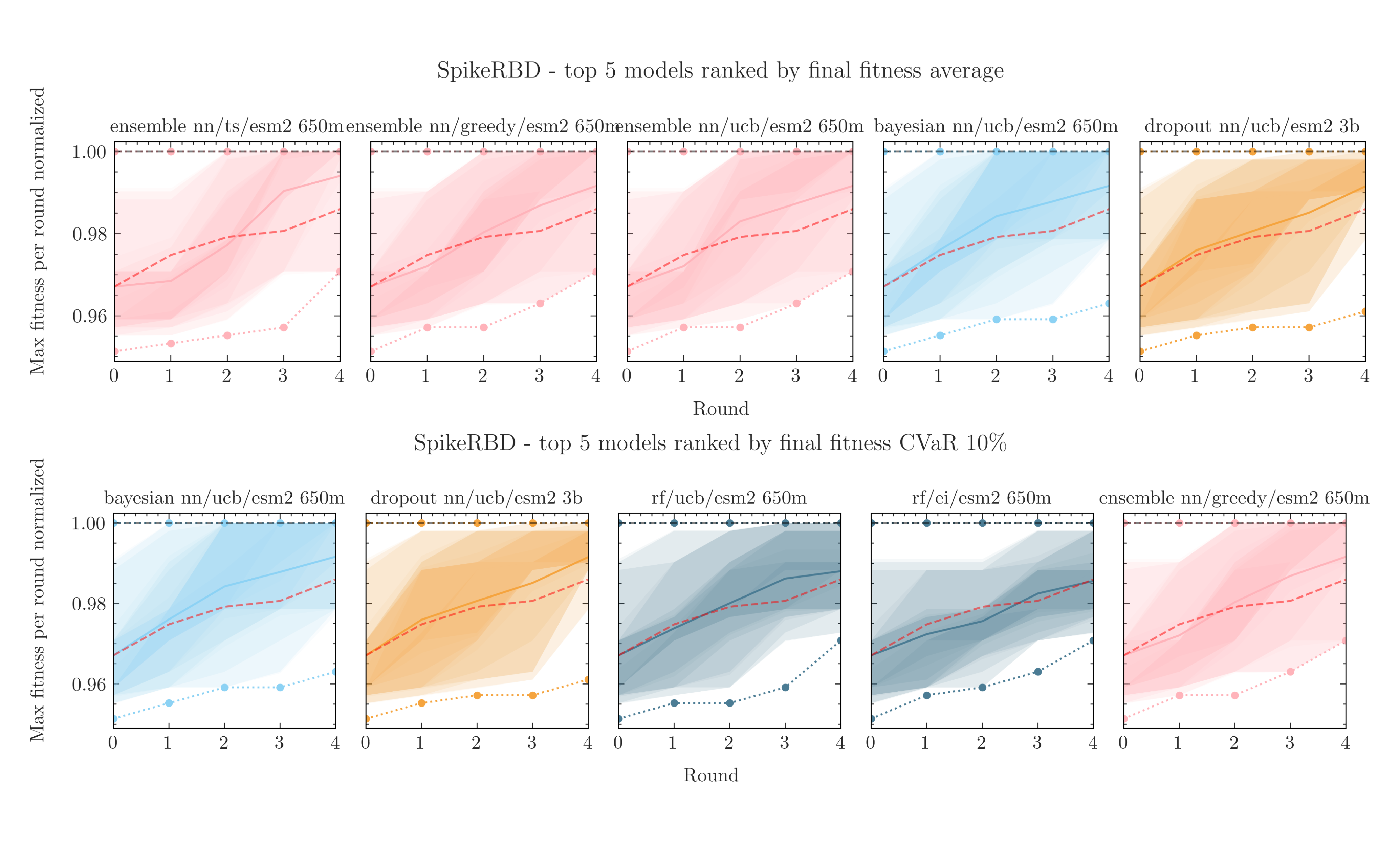}
    \caption{Top 5 models for the average and CVaR-based selection on the final fitness reached - SpikeRBD dataset.}
    \label{fig:hlaa}
\end{figure}

\subsubsection{Dlg4 top models}
\begin{figure}[H]
    \centering
    \includegraphics[width=1\textwidth]{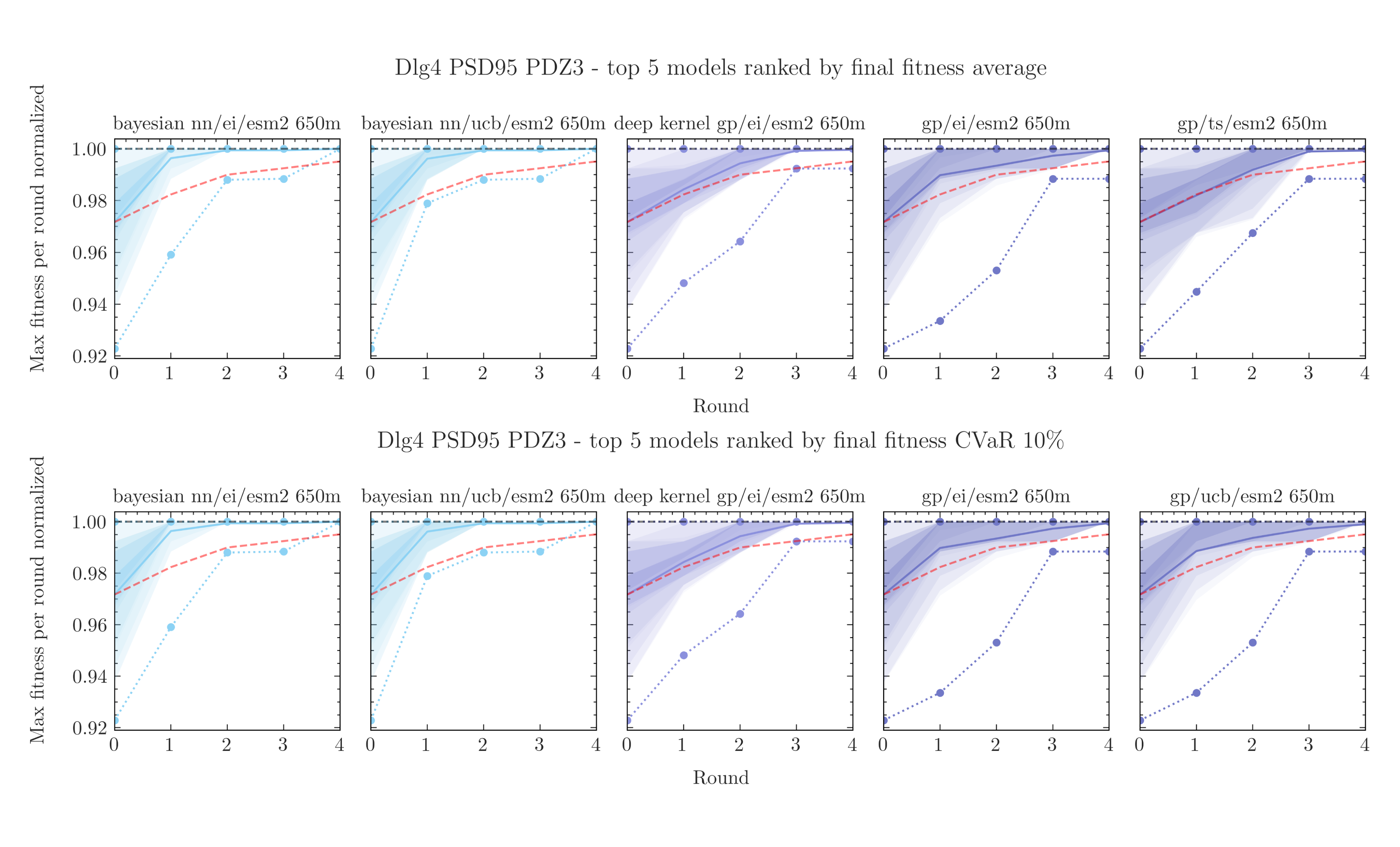}
    \caption{Top 5 models for the average and CVaR-based selection on the final fitness reached - Dlg4 dataset.}
    \label{fig:hlaa}
\end{figure}

\subsubsection{Gcn4 top models}
\begin{figure}[H]
    \centering
    \includegraphics[width=1\textwidth]{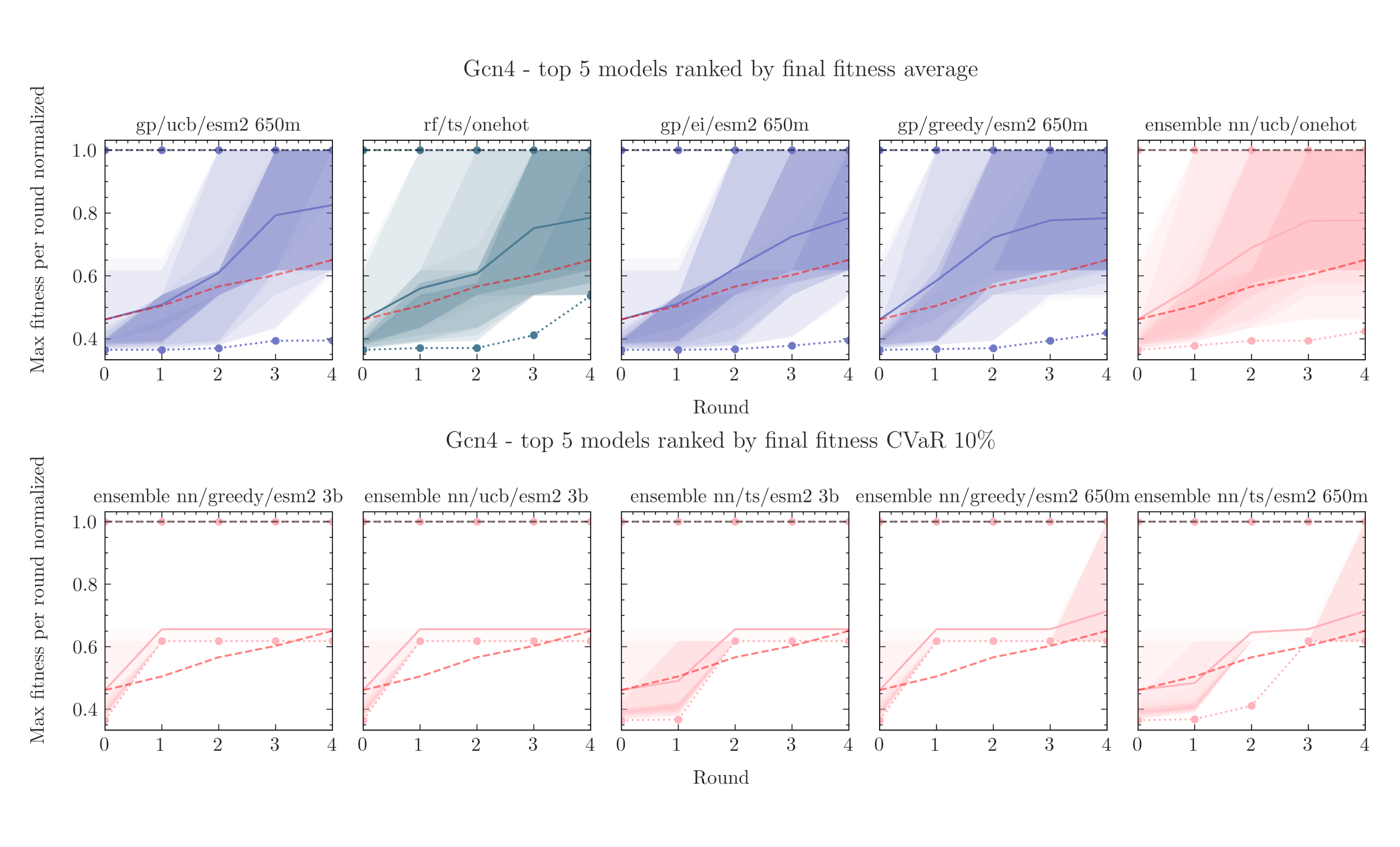}
    \caption{Top 5 models for the average and CVaR-based selection on the final fitness reached - Ccr5 dataset.}
    \label{fig:gcn4}
\end{figure}

\subsubsection{ACE2 top models}
\begin{figure}[H]
    \centering
    \includegraphics[width=1\textwidth]{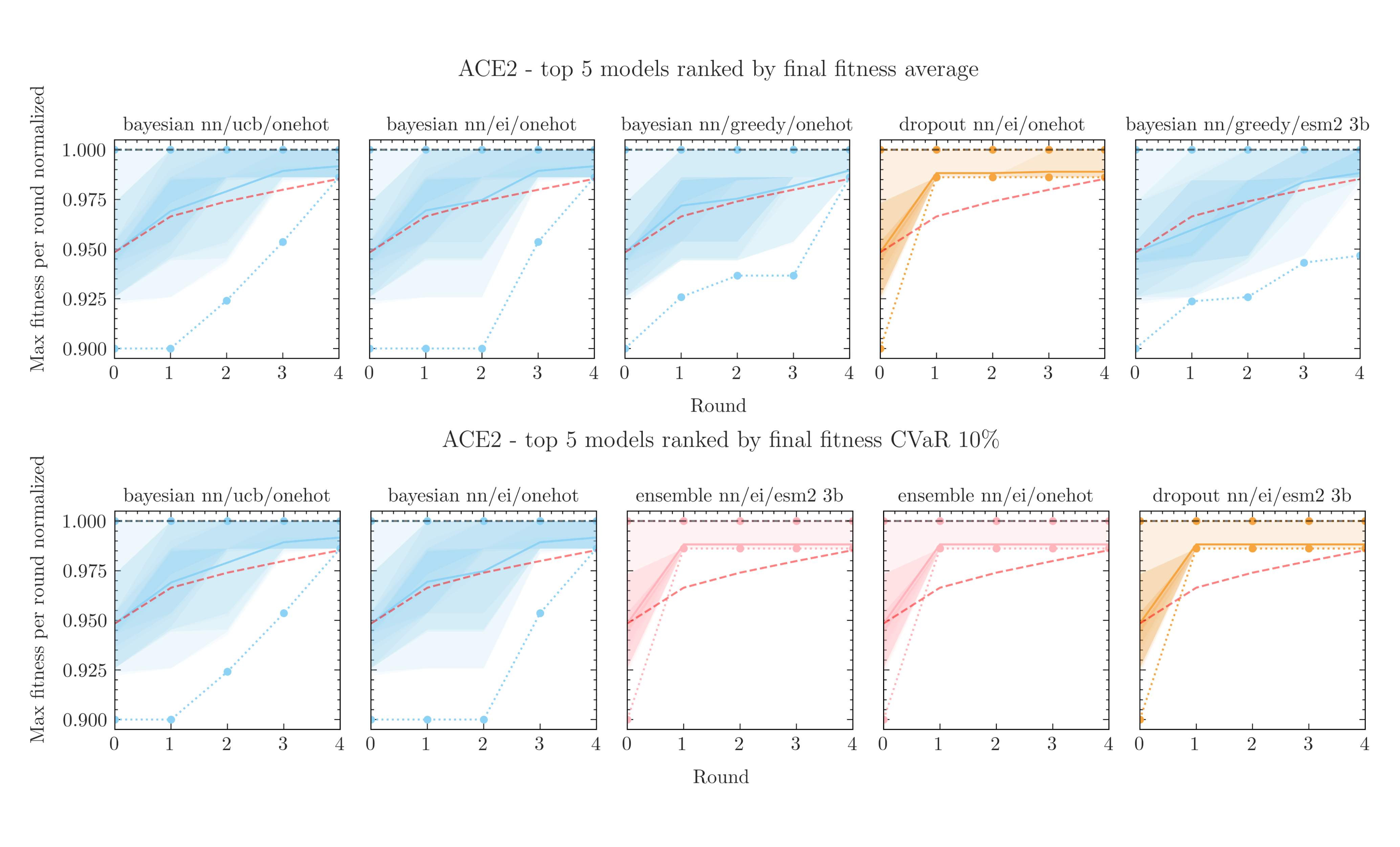}
    \caption{Top 5 models for the average and CVaR-based selection on the final fitness reached - ACE2 dataset.}
    \label{fig:ace2}
\end{figure}

\subsubsection{YAP1 top models}
\begin{figure}[H]
    \centering
    \includegraphics[width=1\textwidth]{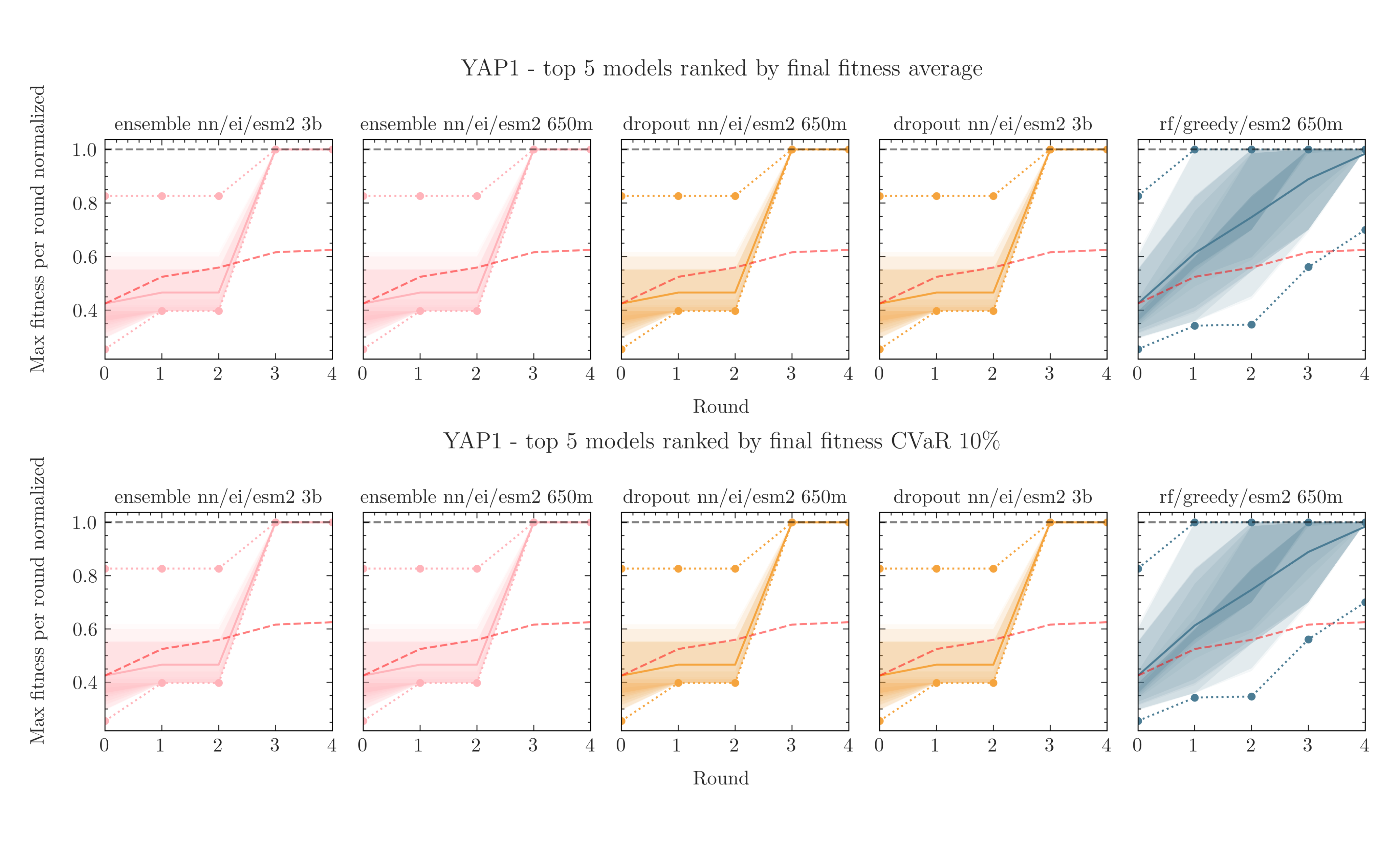}
    \caption{Top 5 models for the average and CVaR-based selection on the final fitness reached - YAP1 dataset.}
    \label{fig:yap1}
\end{figure}

\subsection{Cost-performance Pareto fronts for the GB1 subset dataset}
\begin{figure}[H]
    \centering
    \includegraphics[width=1\textwidth]{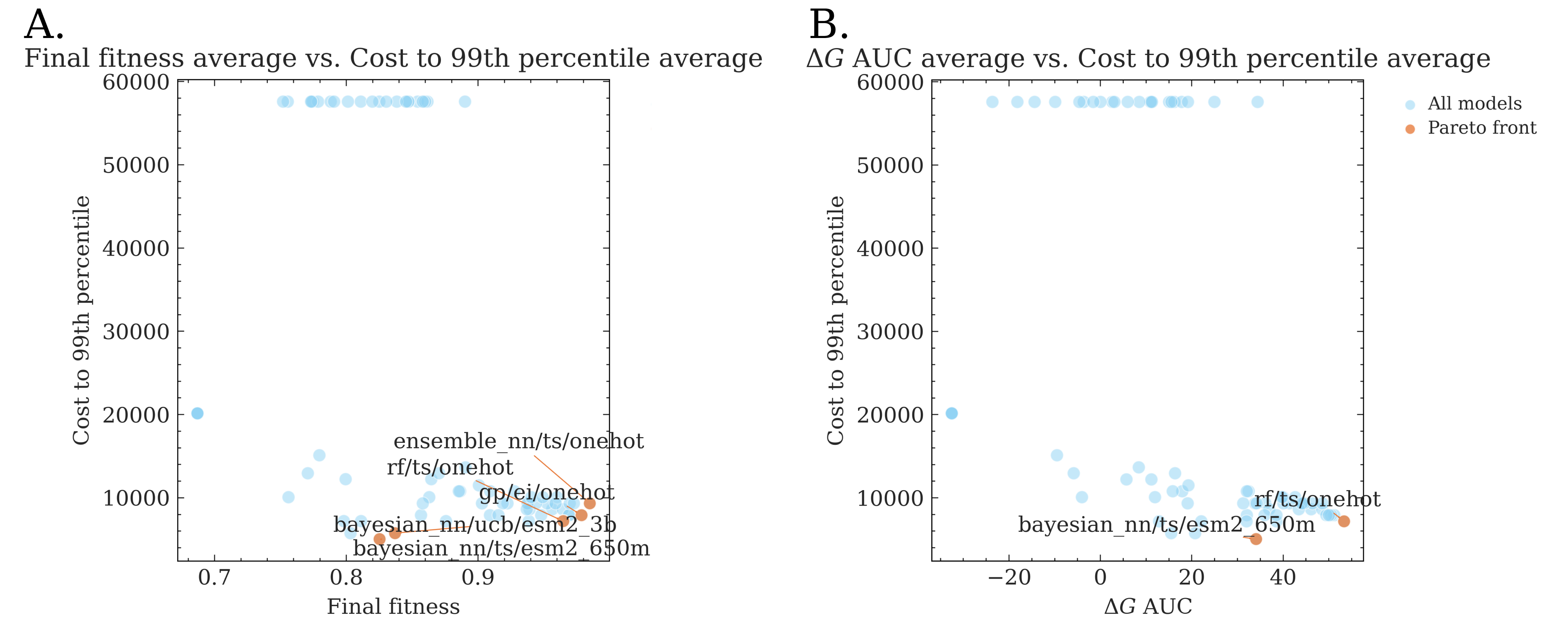}
    \caption{Performance (as the average final fitness reached and average $\Delta{G}\>AUC$) versus costs to reach the 99th fitness percentile. We have highlighted the Pareto-optimal models.}
    \label{fig:suppl - pareto}
\end{figure}

\end{document}